\documentclass[10pt,twocolumn,letterpaper]{article}

\usepackage{cvpr}              

\usepackage{graphicx}
\usepackage{amsmath}
\usepackage{amssymb}
\usepackage{booktabs}
\usepackage[dvipsnames]{xcolor}
\usepackage{multirow}
\usepackage{pifont}

\usepackage[pagebackref,breaklinks,colorlinks]{hyperref}

\usepackage[capitalize]{cleveref}
\crefname{section}{Sec.}{Secs.}
\Crefname{section}{Section}{Sections}
\Crefname{table}{Table}{Tables}
\crefname{table}{Tab.}{Tabs.}

\def\cvprPaperID{2565} 
\def\confName{CVPR}
\def\confYear{2023}

\def\vl{{VL}}
\def\svlcfull{{Structured \vl{} Concepts}}
\def\svlc{{SVLC}}

\newcommand{\cmark}{\ding{51}}%
\newcommand{\xmark}{\ding{55}}%

\def\showcomments{0}
\newcommand\ifcom[3]{
\ifx\showcomments\undefined

\else
  \if\showcomments 1
  \textcolor{#1}{\small\bf\sf [#2: #3]}
  \else
  {}
  \fi
\fi
}

\definecolor{darkgreen}{rgb}{0.0,0.6,0.0}

\definecolor{cyan}{rgb}{0.45,0.87,0.95}
\definecolor{orange}{rgb}{0.95,0.8,0.6}
\definecolor{britishracinggreen}{rgb}{0.0, 0.26, 0.15}
\definecolor{cadmiumgreen}{rgb}{0.0, 0.42, 0.24}

\newcommand\secvspace{\vspace{-0.049cm}}

\newcommand\figvspacetop{\vspace{-0.1cm}}
\newcommand\figvspace{\vspace{-0.3cm}}
\newcommand\tabvspace{\vspace{-0.349cm}}
\newcommand\tabtopvspace{\vspace{-0.2cm}}

\newcommand{\gcol}[1]{{\bf \fontsize{5.5}{42}\selectfont \color{citecolor!80}~(#1)}}
\newcommand{\rcol}[1]{{\bf \fontsize{5.5}{42}\selectfont \color{lightred!180}~(#1)}}
\definecolor{citecolor}{RGB}{34,139,34}
\definecolor{lightred}{RGB}{241,140,142}

\begin{document}


\title{Teaching Structured Vision \& Language Concepts to Vision \& Language Models}

\newcommand{\minisection}[1]{\noindent{\textbf{#1}.}}

\author{
    Sivan Doveh$^{1,2}$,
    Assaf Arbelle$^{1}$,
    Sivan Harary$^{1}$,
    Eli Schwartz$^{1,3}$,
    Roei Herzig$^{1,3}$,\\
    Raja Giryes$^{3}$,
    Rogerio Feris$^{4}$,
    Rameswar Panda$^{4}$,
    Shimon Ullman*$^{2}$,
    Leonid Karlinsky\thanks{Equal contribution}\hspace{-1pt}*$^{4}$\\
    \\
    \tt\small
    $^{1}$IBM Research,
    $^{2}$Weizmann Institute of Science,
    $^{3}$Tel-Aviv University, 
    \tt\small
    $^{4}$MIT-IBM Watson AI Lab
}

\maketitle

\begin{abstract}
\secvspace

Vision and Language (\vl{}) models have demonstrated remarkable zero-shot performance in a variety of tasks.
However, some aspects of complex language understanding still remain a challenge.
We introduce the collective notion of Structured Vision \& Language Concepts (\svlc{}) which includes object attributes,  relations, and states which are present in the text and visible in the image. Recent studies have shown that even the best \vl{} models struggle with \svlc{}. A possible way of fixing this issue is by collecting dedicated datasets for teaching each \svlc{} type, yet this might be expensive and time-consuming. Instead, we propose a more elegant data-driven approach for enhancing \vl{} models' understanding of \svlc{}s that makes more effective use of existing \vl{} pre-training datasets and does not require any additional data. While automatic understanding of image structure still remains largely unsolved, language structure is much better modeled and understood, allowing for its effective utilization in teaching \vl{} models. In this paper, we propose various techniques based on language structure understanding that can be used to manipulate the textual part of off-the-shelf paired \vl{} datasets. \vl{} models trained with the updated data exhibit a significant improvement of up to 15\% in their \svlc{} understanding with only a mild degradation in their zero-shot capabilities both when training from scratch or fine-tuning a pre-trained model. Our code and pretrained models are available at: \url{https://github.com/SivanDoveh/TSVLC}

\end{abstract}

\secvspace
\section{Introduction}
\label{sec:intro}
\secvspace

\begin{figure}[t]
    \figvspacetop
    \centering
    \includegraphics[width=0.35\textwidth]{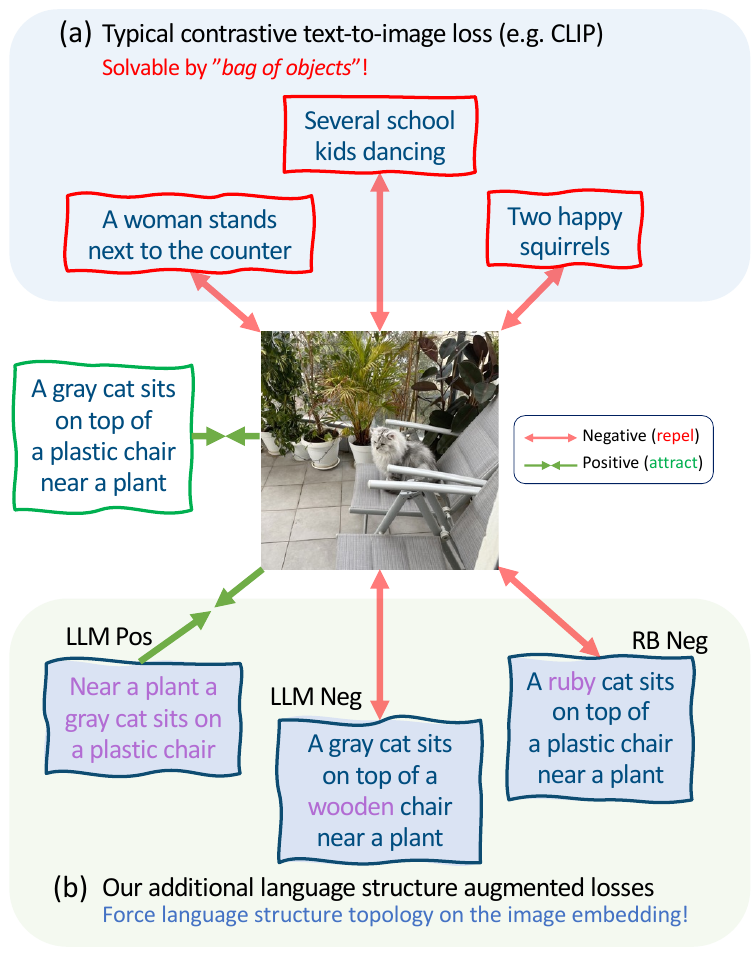}
    \caption{\textbf{Teaching language structure to \vl{} models.} (a) Standard contrastive text-to-image loss (e.g. CLIP \cite{clip}) tends to under-emphasize \svlc{} content of the text, likely due to the random nature of the training batches; (b) We generate modified versions of corresponding texts and use them to add losses to explicitly teach language structure (\svlc{}) to \vl{} models.}
    \label{fig:detailed}
    \figvspace
\end{figure}

%

%
Recent Vision \& Language (\vl) models~\cite{clip,align,albef,blip,cyclip,declip} achieve excellent zero-shot performance with respect to various computer-vision tasks such as detection, classification, segmentation, etc. However, recent studies \cite{winoground,vlc}  
have demonstrated that even the strongest \vl{} models struggle with the compositional understanding of some basic \svlcfull{} (\svlc) such as object attributes, inter-object relations, transitive actions, object states and more. Collecting specialized large scale data to \textit{teach} \vl{} models these missing `skills' is impractical, as finding specialized text-image pairs for each kind and possible value of the different attributes, relations, or states, is both difficult and expensive. 

\begin{figure*}[t]
    \figvspacetop
    \centering
     \includegraphics[width=1.0\textwidth]{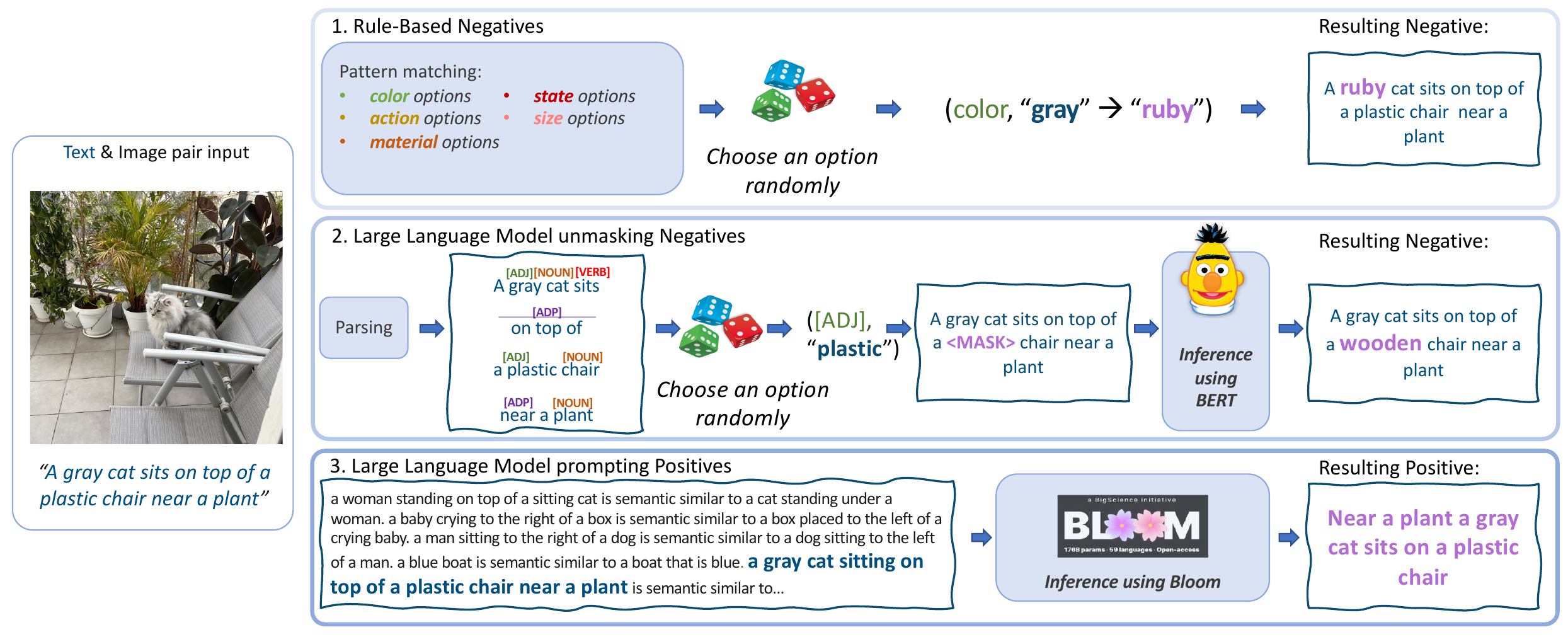} 
    \caption{Teaching structured image understanding to \vl{} models via structured textual data manipulation harnessing the power of language modeling. (1) Generating Rule-Based negative texts (\cref{sec:rb_neg}); (2) Generating negatives using Large Language Model (LLM) unmasking (\cref{sec:auto_neg}); (3) Generating analogies (positives) via LLM prompting (\cref{sec:pos}).}
    \label{fig:intro}
    \figvspace
\end{figure*}
Another important challenge in training \vl{} models with new concepts is catastrophic forgetting, which is a common property to all neural models \cite{kemker:2017,li2016learning,kirkpatrick2017overcoming,castro2018end,ramasesh2020anatomy} and has been explored for \vl{} models in a recent concurrent work \cite{Ding2020CLIPforgetting}. Large \vl{} models such as CLIP \cite{clip} and CyCLIP \cite{cyclip} have exhibited excellent zero-shot learning abilities in many tasks. Therefore, even given a large dataset with new concepts, it is important not to lose these abilities when performing the adaptation to the new data.

In this paper, we propose a way to leverage existing (off-the-shelf) \vl{} pre-training data sources in order to improve the \svlc{} understanding skills of a given model, while at the same time maintaining its zero-shot object recognition accuracy. Naturally, succeeding in this goal would lead to potential improvement w.r.t. \svlc{} understanding in a wide variety of downstream tasks building upon pre-trained \vl{} models, such as zeros-shot detection, segmentation, image generation, and many more.

%


Recent research~\cite{winoground,vlc} has shown that \vl{} models exhibit an `object bias' partially due to the contrastive text-to-image loss used in their pre-training. 
For example, the popular CLIP-loss \cite{clip} is computed over a random batch of text-image pairs sampled from a large-scale and diverse \vl{} dataset with the chance of two images in the same batch containing the same set of objects being very low. 
For such a loss, representing just a 'bag of objects' in each image or text is sufficient for matching the corresponding pairs. 
Intuitively, this leads to the `object bias' where \svlc{}s like attributes, states, and relations are being underrepresented (e.g. having a much smaller amplitude in the resulting feature superposition), consequently causing the aforementioned issues with \svlc{} understanding.

Based on this intuition, we propose a simple data-driven technique that harnesses existing \textit{language} parsing and modeling capabilities to enhance the importance of \svlc{}s in the \vl{} model training losses. For each text in the training batch, we automatically generate alternative negative or positive text by manipulating its content to be opposite or equivalent to the original text. Using the newly generated texts, we explicitly teach \svlc{} to the model via additional losses (see \cref{fig:detailed}) that enforce differentiating between different (original and generated) \svlc{} texts and are no longer satisfiable by the 'bag of objects' representation. 

Towards this end, we propose several techniques for implementing this approach, including (i) rule-based priors based on classical NLP parsing and word substitution vocabulary according to attribute/relation type; (ii) prompting a Large Language Model (LLM) (e.g. \cite{bloom}) for analogous text; (iii) generating different meaning (negative) alternatives by LLM-based unmasking of parsed text entities of different kinds; (iv) combinations of these methods. 

We demonstrate that all these techniques can lead to significant improvements of up to 15\% percent when measuring the \vl{} models' \svlc{} understanding. We verify this on 5 datasets: VG \cite{vg}, HAKE \cite{hake}, VAW \cite{vaw}, SWIG \cite{swig}, and all combined, using the protocol recently proposed in VL-Checklist \cite{vlc}. In addition, we show that the resulting \vl{} models largely 
preserve their zero-shot object recognition performance. For the latter, we also propose a variant of efficient LLM fine-tuning using low-rank residual adapters (LoRA) \cite{lora} adjusted to \vl{} models.  Finally, we show that our framework allows better harnessing of the standard available \vl{} data, e.g. CC3M \cite{cc3m} and LAION \cite{laion}. This is exhibited by the aforementioned gains, both in new \vl{} models trained from scratch, as well as in models fine-tuned using strong available \vl{} models such as CLIP \cite{clip} and CyCLIP \cite{cyclip}. 

\begin{table*}[]
    \centering
    \begin{small}
    \begin{tabular}{ll|cc|lll|lr}

            \toprule
            &&&&\multicolumn{3}{c|}{VL-Checklist}& 21 Zero-Shot&\\
            Dataset&Model&Pre&Arch&Object &Attribute&Relation &Tasks Average&\\
            
            \cmidrule[1.1pt]{1-9}
            
            \multirow{19}{*}{CC3M}&CLIP\cite{clip}&\cmark&Vit-B/32 & 81.58\%&	67.60\%&63.05\%&56.37\%& \textbf{(a)}\\
            \cmidrule{2-8}
            &CLIP + LoRA&\cmark &Vit-B/32& 80.93\%\rcol{-0.66\%}&	66.28\%\rcol{-1.32\%}&	55.52\%\rcol{-7.53\%}&56.41\%\gcol{+0.04\%}&\\
            &CLIP + Ours RB Neg&\cmark &Vit-B/32&	83.89\%\gcol{+2.30\%}&	73.35\%\gcol{+5.75\%}&	75.33\%\gcol{+12.28\%}&54.32\%\rcol{-2.05\%}&\\
            &CLIP + Ours LLM Neg&\cmark&Vit-B/32	&84.44\%\gcol{+2.85\%}&	71.63\%\gcol{+4.03\%}&	74.82\%\gcol{+11.77\%}&55.60\%\rcol{-0.77\%}&\\
            &CLIP + Ours RB+LLM Negs&\cmark&Vit-B/32 &85.09\%\gcol{+3.50\%}&	73.90\%\gcol{+6.30\%}&	78.72\%\gcol{+15.67\%}&54.66\%\rcol{-1.71\%}&\\
            &CLIP + Ours Combined&\cmark&Vit-B/32 &85.00\%\gcol{+3.42\%}&	71.97\%\gcol{+4.37\%}&	68.95\%\gcol{+5.90\%} &54.77\%\rcol{-1.60\%}&\\
            
            \cmidrule[1.1pt]{2-9}
            
&CLIP\cite{clip} &\cmark&Vit-B/16 & 82.91\%&	67.32\%&61.80\%&60.00\%&\textbf{(b)}\\
\cmidrule{2-8}    
&CLIP + Ours RB+LLM Negs &\cmark&Vit-B/16 &85.82\%\gcol{+2.91\%} &	73.92\%\gcol{+6.6\%}&77.40\%\gcol{+15.6\%}&59.37\%\rcol{-0.63\%}&\\
&CLIP + Ours Combined &\cmark&Vit-B/16 &84.75\%\gcol{+1.84\%} & 71.18\%\gcol{+3.86\%}&	69.68\%\gcol{+7.88\%}&59.87\%\rcol{-0.13\%}&\\
            
            \cmidrule[1.1pt]{2-9}

            &CLIP&\xmark&Vit-B/32 &71.17\%&	57.86\%  &	45.20\% &21.96\%&\textbf{(c)}\\
            &CLIP + Ours Combined&\xmark&Vit-B/32 &71.79\%\gcol{+0.62\%}& 63.29\%\gcol{+5.43\%}&	58.13\%\gcol{+12.93\%} &20.96\%\rcol{-1.00\%}&\\

            \cmidrule[1.1pt]{2-9}  
            &CLIP\cite{clip} &\xmark&Vit-B/16 & 64.01\%&	54.27\%&41.57\%&15.49\%&\textbf{(d)}\\
            \cmidrule{2-8}    
            &CLIP + Ours RB+LLM Negs &\xmark&Vit-B/16 &73.11\%\gcol{+9.1\%}&65.32\%\gcol{+11.05}&71.93\%\gcol{+30.36}&20.78\%\gcol{+5.29\%}&\\
            &CLIP + Ours Combined &\xmark&Vit-B/16 &72.99\%\gcol{+8.98\%}&	63.01\%\gcol{+8.74\%}&62.95\%\gcol{+21.38}&20.61\%\gcol{+5.12\%}&\\
            
            \cmidrule[1.1pt]{2-9}   

            &CyCLIP~\cite{cyclip}&\cmark& R50 & 73.49\%&	59.33\%&	53.83\%	&26.00\%&\textbf{(e)}\\
            \cmidrule{2-8} 
            &CyCLIP + LoRA&\cmark& R50 &73.30\%\rcol{-0.19\%}& 	58.89\%\rcol{-0.44\%}&	53.03\%\rcol{-0.80\%}&	26.30\%\gcol{+0.30\%}&\\
            &CyCLIP + Ours Combined&\cmark& R50 & 74.20\%\gcol{+0.71\%}&	63.52\%\gcol{+4.20\%}&	59.47\%\gcol{+5.63\%}&	26.31\%\gcol{+0.31\%}&\\
            
            \cmidrule[1.1pt]{2-9}  
            
            &CyCLIP&\xmark& R50 &69.41\%&	57.59\% 	 &	53.70\% 	&21.02\%&\textbf{(f)}\\
            &CyCLIP + Ours Combined&\xmark& R50 &71.50\%\gcol{+2.09\%}&	65.69\%\gcol{+8.10\%}&	70.20\%\gcol{+16.50\%}&20.44\%\rcol{-0.42\%}&\\
            
            \midrule[1.5pt]  
            
            \multirow{3}{*}{LAION}&CLIP~\cite{clip}&\cmark&Vit-B/32 &81.58\% &67.6\%&63.05\%&56.37\%&\textbf{(g)}\\
            \cmidrule{2-8}
            &CLIP + LoRA&\cmark&Vit-B/32 &82.18\%\gcol{+0.60\%} &	68.48\%\gcol{+0.88\%}& 	62.72\%\rcol{-0.33\%}&57.15\%\gcol{+0.78\%}&\\
            &CLIP + Ours Combined&\cmark&Vit-B/32& 82.54\%\gcol{+0.96\%} &	69.64\%\gcol{+2.04\%}& 	66.05\%\gcol{+3.00\%}&56.71\%\gcol{+0.34\%}&\\

            \bottomrule

    \end{tabular}
    
    \end{small}
    \caption{\textbf{VL-Checklist and Zero-Shot classification evaluation (ImageNet + 20 datasets)}. Finetuned models (for 5 epochs as detailed in \cref{sec:ft}, starting from officially released CLIP \cite{clip} and CyCLIP \cite{cyclip} weights) are marked with \cmark in the Pre-trained (\textit{Pre}) column, while models trained from scratch (for 10 epochs) are marked with \xmark. The \textcolor{OliveGreen}{gains} and \textcolor{red}{losses} of our approach (\textbf{+Ours}) are in color and are computed w.r.t. to corresponding baselines in each section. CLIP/CyCLIP + LoRA indicate finetuning in the same way and on the same data, but without our approach. Finetuning without LoRA on the same data yields significantly worse performance in all metrics. Sections are separated by \textbf{bold} horizontal lines. \textbf{(a)} \textit{CC3M fine-tuning - CLIP - Vit-B/32}: we significantly improve the \svlc{} understanding, observing only small ZS performance drops, $0.77\%$; \textbf{(b)} \textit{CC3M fine-tuning - CLIP - Vit-B/16}: we can observe similar improvements as in (a), with an even smaller impact on ZS performance; \textbf{(c)} \textit{CC3M from scratch - CLIP - Vit-B/32}: we significantly improve \svlc{} understanding with only a small ($1\%$) decrease in ZS performance; \textbf{(d)} \textit{CC3M from scratch - CLIP - Vit-B/16}: compared to (c), even greater \svlc{} understanding improvement is observed (up to $30.36\%$), at no cost in ZS performance, and even improvement of over $5\%$; \textbf{(e)} \textit{CC3M fine-tuning - CyCLIP}: we use CyCLIP original code (with LoRA integration) and losses, as can be seen - adding our techniques improves CyCLIP \svlc{} performance considerably without sacrificing ZS performance;  \textbf{(f)} \textit{CC3M from scratch - CyCLIP}: observing even largergains in \svlc{} understanding compared to (e) (up to $16.5\%$), with small reduction in ZS performance of $0.42\%$; \textbf{(g)} \textit{LAION fine-tuning - CLIP - Vit-B/32}: we improve \svlc{} understanding without any decrease in ZS performance.}
    \label{tab:main_res_vl}
    \tabvspace
\end{table*}

To summarize, we offer the following contributions: 
(i)~We propose a data-driven approach for better harnessing the standard available \vl{} data to improve \vl{} models' \svlc{} understanding skills, such as understanding object attributes, inter-object relations, transitive actions, object states, and more, without sacrificing zero-shot object recognition performance; 
(ii) More specifically, we propose to leverage the well-understood and well-modeled structure of language, through classical NLP parsing and/or use of the modern pre-trained LLMs, for manipulating the text part of the standard \vl{} paired datasets to regularize \vl{} training and teach \svlc{} understanding to \vl{} models. 
(iii) We further propose an adaptation of efficient LLM fine-tuning technique of \cite{lora} for fine-tuning \vl{} models, allowing for only minimal reduction in zero-shot object recognition performance after fine-tuning, while still obtaining the aforementioned \svlc{} understanding gains. 
(iv) Empirically, for the popular CLIP \cite{clip} and its most recent extension CyCLIP \cite{cyclip}, we demonstrate \svlc{} understanding average improvements of up to $13\%$ when training from scratch, and $15\%$ when fine-tuning from a pretrained model.

\secvspace
\section{Related Work}
\label{sec:rw}
\secvspace
\minisection{Vision-language (VL) Models} (\eg, CLIP~\cite{clip} and ALIGN~\cite{align}) show significant advances in diverse zero-shot downstream tasks. They are pre-trained using contrastive image-text alignment on a large-scale noisy dataset of text-image pairs collected from the web. Several methods~\cite{tan2019lxmert,chen2020uniter,li2020oscar} additionally employ off-the-shelf object detectors to extract region features. In order to relax this limitation, some methods~\cite{kim2021vilt,align,yang2022vision,blip} propose to use cross-attention layers with self-supervised learning objectives including image-text matching and masked/autoregressive language modeling. 
BLIP~\cite{blip} generates synthetic captions from the language modeling head and filters noisy captions based on the image-text matching score. Recently, there have been attempts to learn finer-level alignment and relations between image and text~\cite{cyclip,yao2021filip,furst2021cloob,declip,gao2022pyramidclip}. FILIP proposes fine-grained contrastive learning to maximize the token-wise similarity between visual and textual tokens. CyClip~\cite{cyclip} imposes additional geometrical consistency on the image and text embeddings. DeCLIP~\cite{declip} introduces additional positives from the nearest neighbors. While these methods improve image-text retrieval tasks on the existing benchmarks, such as ImageNet\cite{russakovsky2015imagenet} and  MS-COCO~\cite{lin2014microsoft}, recent studies such as VL-CheckList~\cite{zhao2022vl} and the Winoground Challenge \cite{winoground}, show that these models cannot distinguish fine-grained language details or understand structured concepts (SVLCs) such as object attributes and relations. In this paper, we focus on the latter and propose orthogonal data-driven techniques that have the potential to improve the \svlc{} understanding for all \vl{} models.

\minisection{Learning Structured Representations} A full understanding of the semantics of rich visual scenes requires the ability to understand visual concepts, such as detecting individual entities and reasoning about their interactions and attributes. Structured representations have played an important role in achieving this goal, having been successfully applied to a wide range of computer vision applications: vision and language~\cite{Chen2020UNITERUI,Li2019VisualBERTAS,li2020oscar,Tan2019LXMERTLC}, scene graphs~\cite{sg_generation_msg_pass,herzig2018mapping,referential_relationships,Jerbi2020LearningOD,raboh2020dsg}, relational reasoning~\cite{baradel2018object,battaglia2018relational}, human-object interactions~\cite{Gao2020DRGDR,Kato2018CompositionalLF,Xu2019LearningTD}, action recognition~\cite{avraham2022svit,arnab2021unified,materzynska2019something,herzig2022orvit,herzig2019stag,ji2019action,Wang_videogcnECCV2018}, and even image \& video generation from graphs~\cite{2020ActionGraphs,herzig2019canonical,johnson2018image}. 
However, most of these works rely on detailed, manually curated, supervision, often involving annotation of location information and structural details, which are very expensive to collect and scale, resulting in limited-size or synthetic data sources for training. In contrast, in our work, we focus on methods for teaching \svlc{} understanding to large \vl{} models while only leveraging the available large-scale noisy \vl{} data sources collected from the web without any use of expensive manual curation. 



\minisection{Data Augmentation} Augmentation plays a key role in many computer vision applications~\cite{sohn2020fixmatch,chen2020simple}. Several advanced image augmentation methods (CutMix~\cite{yun2019cutmix}, mixup~\cite{zhang2017mixup}, AutoAugment~\cite{cubuk2018autoaugment}, RandAugment~\cite{cubuk2020randaugment}, etc) have been proposed and greatly improved computer vision task performance.  Text augmentation has been tackled trough back-translation~\cite{xie2020unsupervised}, word and frame-semantic embedding augmentations~\cite{wang2015s}, word replacement~\cite{zhang2015character}, random word insertion/deletion/swap~\cite{wei2019eda}, or using a text generative model~\cite{kumar2020data} in diverse NLP applications. In VL tasks, previous work explores machine translation between different languages~\cite{kim2020mule,burns2020learning}, generating synthetic captions~\cite{blip}, adversarial/synthetic data augmentation for VQA~\cite{tang2020semantic,ray2019sunny} or mixup for VL~\cite{hao2022mixgen}. 
We focus on leveraging the well-understood and modeled language structure for manipulating text in a way that explicitly targets teaching \svlc{} semantics to \vl{} models. To the best of our knowledge, this has not been attempted before.
\secvspace
\section{Method}
\label{sec:method}
\secvspace
In this section, we discuss the proposed framework for improving \svlc{} performance of \vl{} models using already available \vl{} data. \cref{sec:teaching_svlc} and \cref{sec:losses} present our main techniques for teaching \svlc{} to \vl{} models. These approaches can be effectively applied both for fine-tuning existing strong \vl{} Pre-trained models, as well as for training \vl{} models from scratch. In both cases, they exhibit significantly improved \svlc{} performance as demonstrated in our experiments in \cref{sec:experiments}. \cref{sec:ft} presents our strategy for fine-tuning \vl{} models on \svlc{}-enhanced \vl{} data, while at the same time being parameter efficient and significantly reducing forgetting, thus, maintaining the \vl{} model Zero-Shot (ZS) performance. Qualitative examples showing improvements attained by our proposed approach, as well as some examples of failure cases, are provided in the supplementary material.


\secvspace
\subsection{Teaching \svlc{} Using NLP and LLMs}\label{sec:teaching_svlc}
\secvspace
In this section, we present several ways in which the data of existing \vl{} pre-training paired datasets can be enhanced to emphasize \svlc{} in the texts, and teaching them to the \vl{} model. 
We propose two kinds of data enhancements - generating negative and positive text alternatives. When generating negative examples, only one word of the sentence is changed such that the semantic meaning of the sentence changes. We propose two methods for the generation of the negatives: (i) rule-based (\cref{sec:rb_neg}); and (ii) LLM-based (\cref{sec:auto_neg}). 
Positive alternatives are generated as sentences with semantically similar meanings, but different wording (\cref{sec:pos}). We then present the losses which properly take these two types of generated textual data into account during training in \cref{sec:losses}.
            
            
    

\secvspace
\subsubsection{Generating Rule-Based (RB) Negatives}\label{sec:rb_neg}
\secvspace
One simple yet effective method for negative text generation is using a collection of pre-defined language rules which match and replace words of a specific entity type or a pattern, such as color, material, size, etc. 
This method is especially useful when one has prior knowledge of a specific aspect of the language that needs to be taught. For example, if we know that our model lacks the ability to understand the colors of objects, we can easily create a rule for detecting and replacing color words in the text, for generating negative text that does not correspond to its paired image. 
To employ the generation of the rule-based negative, for each taught \svlc{} we define a list of words belonging to its characteristic. We then scan the \vl{} data texts searching for the words within these lists. If a word is located, we simply replace it with a randomly selected word from the same list to generate a negative pair. For example, applying the color-rule to a sentence: ``A big \textbf{brown} dog" can lead to ``A big \textbf{yellow} dog". If a text has multiple candidates of words to be replaced, one of them is chosen randomly. 
We perform this process multiple times for the full list of \svlc{} characteristics of interest such as color, size, material, spatial relations, etc. These generated negative texts are \svlc{} specific and differ from the original text in only one word. For a detailed description and more examples of the RB negative generation, please refer to the supplementary material. 

\secvspace
\subsubsection{LLM-based Negative Generation via Unmasking}\label{sec:auto_neg}
\secvspace
A natural extension of rule-based negatives technique is the generation of negatives using Large Language Models (LLMs) unmasking. Recent LLMs are explicitly trained in a self-supervised manner with the objective of ``unmasking" parts of the text. 
Given a sentence with one missing word, models such as BERT~\cite{bert}, can suggest multiple words that fit the context of the sentence. Using this useful property of LLMs, we can therefore automatically create plausible negative examples without the need for prior knowledge of the \svlc{} characteristics of interest. In order to focus the randomly selected masked words to be likely to belong to \svlc{}s of interest, we use common NLP parsing techniques (such as spacy~\cite{spacy}) to parse the sentence into its components such as nouns, verbs, adjectives, adverbs, etc. We then randomly choose a type of sentence part and a word belonging to this part type, mask out the selected word and replace it with one of the options suggested by the LLM's unmasking. These negative examples, when used properly in the loss function (\cref{eq:neg}) focus the network on the important details that affect the \svlc{} understanding.  As we show in \cref{sec:experiments}, this method is extremely useful and can significantly improve the \vl{} model's understanding of different \svlc{}s. Further details and examples of the LLM negative generation are provided in the supplementary material.

\secvspace
\subsubsection{Generating Text Analogies via LLM Prompting}\label{sec:pos}
\secvspace
While the goal of the negative text generation (\cref{sec:rb_neg} and \cref{sec:auto_neg}) was to make minor perturbations to a given text such that the meaning changes, the goal here is exactly the opposite. We would like to make major changes to the text, while still keeping the same semantic meaning. 
For example ``A woman standing left to a sitting cat" and ``A cat sitting to the right of a standing woman" are two very different texts describing the exact same scene. One effective way to generate such semantically similar texts is by prompting the foundational LLMs. Specifically, we use the open access BLOOM~\cite{bloom} model. In the spirit of recently popular in-context learning~\cite{incotext}, we present the model with a textual prompt with examples of semantically similar texts (see \cref{fig:intro}). We then append the current image caption and retrieve the BLOOMs prediction of a semantically similar text. For a detailed description, we refer the reader to the supplementary material. 



\secvspace
\subsection{Losses}\label{sec:losses}
\secvspace

All of our evaluated models (CLIP \cite{clip} and CyCLIP \cite{cyclip}) admit a text \& image pair $(T,I)$ and are comprised of two parts: (i) image encoder $e_I = \mathcal{E}_I(I)$; (ii) text encoder $e_T = \mathcal{E}_T(T)$. In this notation, the text-to-image similarity score is therefore computed as:
\begin{equation}
    \mathcal{S}(T,I) = \exp\left(\frac{\tau e_T^Te_I}{||e_T||^2||e_I||^2}\right),
\end{equation}
where  $\tau$ is a learned temperature parameter.



\noindent\textbf{Contrastive Loss.} As most contemporary \vl{} models, we employ the contrastive CLIP-loss \cite{clip} as one of our losses for each batch  $\mathcal{B}$.
\begin{equation}\label{eq:contrastive}
    \mathcal{L}_{cont} = \sum_{i}log\left(\frac{{S}(T_i,I_i) } {\sum_{j}{{S}(T_i,I_j)}}\right) +log\left(\frac{{S}(T_i,I_i) } {\sum_{k}{{S}(T_k,I_i)}}\right).
\end{equation}



\noindent\textbf{Negatives Loss.} In our ablation study in \cref{sec:ablations}, we show that for a given text $T_i$ simply adding the corresponding generated negative text $T_i^{neg}$ to the contrastive loss $\mathcal{L}_{cont}$ is much less effective than having a separate loss individually attending to the similarity difference of $T_i$ and $T_i^{neg}$ w.r.t. the image $I_i$ corresponding to $T_i$. We, therefore, employ the following \textit{negatives loss}:
\begin{equation}
\label{eq:neg}
    \mathcal{L}_{neg} = \sum_{i}-log\left(\frac{\mathcal{S}(T_i,I_i)}{\mathcal{S}(T_i,I_i) + \mathcal{S}(T_i^{neg},I_i)}\right).
\end{equation}

\noindent\textbf{Analogy Loss.} For the generated analogy texts $T_i^{sim}$ produced from the text $T_i$ which corresponds to image $I_i$, we employ the combination of the two following losses:
\begin{equation}
    \mathcal{L}_{sim}^{text} = \sum_{i}{-log\left(\frac{{\mathcal{S}(T_i^{sim},T_i)}}{\sum_{j}{{\mathcal{S}(T_i^{sim},T_j)}}}\right)},
\end{equation}
where with some abuse of notation, $\mathcal{S}(T_1,T_2)$ denotes the exponent cosine similarity between text $T_1$ and text $T_2$ text-embeddings, and:
\begin{equation}
    \mathcal{L}_{sim}^{img} =
    \sum_{i}{-log\left(\frac{{\mathcal{S}(T_i^{sim},I_i)}}{\sum_{j}{{\mathcal{S}(T_i^{sim},I_j)}}}\right)},
\end{equation}
%
%
which simply corresponds to the second summand of $\mathcal{L}_{cont}$ in \cref{eq:contrastive}, with replacing $T_i$ by $T_i^{sim}$.

Our final loss is, therefore:
\begin{equation}
    \mathcal{L}=\mathcal{L}_{cont} + \alpha \cdot \mathcal{L}_{neg} + \beta \cdot (\mathcal{L}_{sim}^{text} + \mathcal{L}_{sim}^{img}).
    \label{eq:final_loss}
\end{equation}

\secvspace
\subsection{Fine-tuning Pre-trained \vl{} Models}\label{sec:ft}
\secvspace
Each of the $\mathcal{E}_T$ and $\mathcal{E}_I$ networks is comprised of a mix of non-parametric functions
and two types of parametric functions: linear layers
and embedding layers. 
Roughly, each of those functions, $\mathcal{F}^{lin}_k(x)$ and $\mathcal{F}^{emb}_k(x)$ where $k$ is layer index,
is parameterized by a weight matrix $\mathcal{W}_k$ so that:
\begin{align}
    &\mathcal{F}^{lin}_k(x)=\mathcal{W}_k \cdot x \\
    &\mathcal{F}^{emb}_k(x)=EMB(x;\mathcal{W}_k)
\end{align}
where $EMB$ is the embedding operator assuming $x$ is a stream of integers and picking the respective columns of $\mathcal{W}_k$. 
Following the idea proposed in LoRA \cite{lora} for efficient LLM fine-tuning using low-rank residual adapters, when adapting a pre-trained \vl{} model $\mathcal{M}=(\mathcal{E}_T,\mathcal{E}_I)$ we parameterize the adapted weights $\mathcal{W}_k^*$ of the model $\mathcal{M}^*$ fine-tuned from $\mathcal{M}$ as:
\begin{equation}
    \mathcal{W}_k^* = \mathcal{W}_k + \mathcal{A}_k \cdot \mathcal{B}_k
\end{equation}
where for $\mathcal{W}_k$ of size $m \times l$, $\mathcal{A}_k$ and $\mathcal{B}_k$ are rank-$r$ matrices of sizes $m \times r$ and $r \times l$ respectively. These low-rank residual adapters can be applied efficiently as:
\begin{align}
    &\mathcal{F}^{*,lin}_k(x)=\mathcal{F}^{lin}_k(x) + \mathcal{A}_k \cdot (\mathcal{B}_k \cdot x) \\
    &\mathcal{F}^{*,emb}_k(x)=\mathcal{F}^{emb}_k(x) + \mathcal{A}_k \cdot EMB(x;\mathcal{B}_k)
\end{align}
During the fine-tuning of $\mathcal{M}^*$, we freeze all the base model $\mathcal{M}$ parameters $\forall k,\{\mathcal{W}_k\}$ and only the LoRA adapters $\forall k,\{(\mathcal{A}_k,\mathcal{B}_k)\}$ are being learned. In the above notation we disregard possible bias terms of the linear functions, if they are present, since we keep them frozen too.

There are several interesting things to note about the proposed architecture: (i) as opposed to \cite{vladapter}, who evaluated the use of efficient LLM fine-tuning techniques for \vl{} models adaptation, we add our LoRA adapters everywhere, i.e to all layers of both the text and image encoders, and not only to the text encoder/decoder as done in \cite{vladapter}; (ii) as opposed to \cite{sidetuning} who attach a small side network only to the \textit{output} of the adapted model, our LoRA adapters are added to all the parametric functions inside the model and affect all the intermediate computations; (iii) same as noted in the original \cite{lora} paper, at inference all the LoRA adapters can be folded back into the original weight matrices by simple summation, thus returning the number of total parameters to be the same as in the original model and hence have the same inference speed; (iv) with rank $r$ kept low, the number of extra parameters added by all the LoRA adapters can be very low
making adaptation fast and efficient; finally, (v) such form of fine low-rank adaptation allows for significantly mitigating the zero-shot performance forgetting effects as demonstrated in our results and explored in the corresponding ablation (\cref{sec:ablations}). 

\secvspace
\section{Experiments}\label{sec:experiments}

\secvspace
\subsection{Datasets}
\secvspace

We train the model using common Image-Text pair datasets, namely Conceptual Captions 3M~\cite{cc3m} and LAION 400M~\cite{laion} and test using the VL-Checklist~\cite{vlc} datasets, which will be described below. 
\\
\textbf{Conceptual Captions 3M (CC3M)~\cite{cc3m}} is a dataset of three million image-text pairs automatically crawled from the internet where image descriptions are harvested from Alt-text attributes and then processed and filtered to create relatively clean descriptions. 
\\
\textbf{LAION 400M~\cite{laion}}~ is a very large scale image-text pair dataset which, similarly to CC3M  has been automatically harvested from the internet. One major difference between the two, apart from the size, is that LAION examples have been filtered using the pretrained CLIP model, such that the CLIP image-text similarity is high by design. Recent re-implementations of the original CLIP paper have succsefly used LAION 400M to reproduce similar capabilites as the original CLIP model. 
\\
\begin{table*}[h!]
    \centering
    \begin{small}
    \begin{tabular}{l|lll|l}

            \toprule
            &\multicolumn{3}{c|}{VL-Checklist}& 21 Zero-Shot Tasks\\
            &Object &Attribute&Relation & Average \\
            \midrule            
            CLIP\cite{clip} & 81.58\%&	67.60\%&63.05\%&56.37\%\\
            \midrule
CLIP +LoRA & 80.93\%	(\textcolor{red}{-0.66\%})&	66.28\%	(\textcolor{red}{-1.32\%})&	55.52\%	(\textcolor{red}{-7.53\%})&56.41\%(\textcolor{OliveGreen}{+0.04\%})\\
 w/o Neg Loss &82.27\% 	(\textcolor{OliveGreen}{+0.69\% }) &	67.58\% 	(\textcolor{red}{-0.02\% })&	55.17\% 	(\textcolor{red}{-7.88\% })&	55.37\% 	(\textcolor{red}{-1.00\% })\\
Ours RB+LLM Negs &85.09\% (\textcolor{OliveGreen}{+3.50\%})&	73.90\%	(\textcolor{OliveGreen}{+6.30\%})&	78.72\%	(\textcolor{OliveGreen}{+15.67\%})&54.66\% (\textcolor{red}{-1.71\%)})\\

            \bottomrule
    \end{tabular}
    
    \end{small}
    \caption{\textbf{Ablation study - separate Negative Losses (\cref{sec:losses}) vs adding negatives to contrastive loss.}}
    \label{tab:ablation-neg}
    \tabvspace
\end{table*}
\textbf{VL-Checklist\cite{vlc}} combines images and annotations from the Visual Genome~\cite{vg}, SWiG~\cite{swig}, VAW~\cite{vaw}, and HAKE~\cite{hake} datasets. It is processed such that each image is annotated with two captions, one positive and one negative. The positive caption originates in its source dataset and corresponds to the image. 
The negative caption is constructed from the positive caption so that only one word, corresponding to the tested \svlc{} of interest, is changed to negate the \svlc{} (\eg, color, size, material, etc.). VL-Checklist evaluates three main types of \vl{} concepts further subdivided into 7 types of \svlc{}s total: (1) Object: its spatial location and size, (2) Attribute: color, material, size, state, and action, and (3) Relation: spatial or action relation between two objects. In the following sections we report results on a combined VL-Checklist dataset. The results on individual comprising datasets (Visual Genome~\cite{vg}, SWiG~\cite{swig}, VAW~\cite{vaw}, and HAKE~\cite{hake}) are provided in the supplementary material.
\\
\textbf{Zero-Shot Classification} We evaluated our method on 21 classification dataset using the Zero-Shot classification protocol described in the ELEVATER Image Classification Toolkit~\cite{elevater}. 
The evaluation inclueds 21 different datasets, including common classification datasets such as ImageNet~\cite{russakovsky2015imagenet}, CIFAR100~\cite{cifar100}, EuroSat~\cite{eurosat}, and others. in Tables~\ref{tab:main_res_vl}-\ref{tab:ablation_comp_att_rel} we report the average results over the 21 tasks.

\secvspace
\subsection{Implementation Details}
\secvspace
For CLIP we use the ML-Foundation Open-CLIP repository~\cite{openclip} and for CyCLIP \cite{cyclip} we use its original code repository, which is also based on Open-CLIP. In most experiments, unless stated otherwise, we train our model for five epochs on 4 V100 NVIDIA GPUs, with a total batch size of 512. When starting from a pre-trained model, we use rank $4$ LoRA adapters (\cref{sec:ft}) and the learning rate is set to $5E-6$. When training from scratch, for CLIP we use the default parameters set in the open-CLIP library, and for CyCLIP the defaults of its original implementation. For all CLIP experiments, we use VIT/32-B as the model architecture and ResNet-50 when training CyCLIP (following \cite{cyclip}). When fine-tuning we initialize with the original model weights released by the respective authors. In all experiments involving a combination of CyCLIP \cite{cyclip} and our method (in \cref{sec:exp-ft-vl-models}-\ref{subsec:fromscratch}), in addition to \cref{eq:final_loss} loss we also employ all the extra losses proposed in CyCLIP \cite{cyclip}.

\secvspace
\subsection{Baselines}
\secvspace
We compare our method to two strong baselines under several configurations. The first is the CLIP~\cite{clip} OpenAI pretrained model trained on 400M image-text pairs which achieves high ZS performance. The second is the very recent CyCLIP~\cite{cyclip} method which, similarly to us, improves over the original CLIP loss. For a fair comparison all methods use the same network architecture and the same initialization from the OpenAI pretrained model.  For the CyCLIP baseline we continue training from the pretrained initialization using LoRA and the CyCLIP losses. As CyCLIP losses are orthogonal to ours, we also show a unified version of the two methods (CyCLIP + Ours Combined).

\begin{table*}[h!]
    \tabtopvspace
    \centering
    \begin{small}
    \begin{tabular}{l|ccccc|cc|c}
            
            \toprule
            &\multicolumn{5}{c|}{Attribute}&\multicolumn{2}{c|}{Relation}&{21 Zero-Shot Tasks}\\
           Method & Color &	Material &	Size &	State &	 Action	& Action &	Spatial & Average\\
           \midrule
Ours Pos & 72.35\% &	69.25\% &	69.80\% &	59.35\% &	66.08\%& 70.97\% &	39.20\%& \textbf{55.37\%} \\
Ours RB Neg &	78.45\% &	83.20\% &	69.50\% &	\textbf{65.95\%} &	69.66\% &	75.97\% &	74.70\% &54.66\%\\
Ours LLM Neg	&	76.00\% &	79.70\% &	72.75\% &	61.35\% &	68.36\% &	77.23\% &	72.40\% &55.05\%\\
Ours RB+LLM Negs&\textbf{79.25\%} &	\textbf{84.25\%} &	72.15\% &	64.05\% &	\textbf{69.82\%} &	\textbf{79.03\%} &	\textbf{78.40\% }&54.66\%\%\\

Ours Combined&77.45\% &	77.35\% &	\textbf{73.35\%} &	62.30\% &	69.39\% &	74.70\% &	63.20\% &54.77\%\%\\

            \bottomrule
    \end{tabular}
    
    \end{small}
    \caption{\textbf{Ablation study - component analysis.} Detailed evaluation on \textit{Attribute} \& \textit{Relation} \svlc{}s.}
    \label{tab:ablation_comp_att_rel}
    \tabvspace
\end{table*}
\begin{figure}[]
    \figvspacetop
    \centering
    \includegraphics[width=0.35\textwidth]{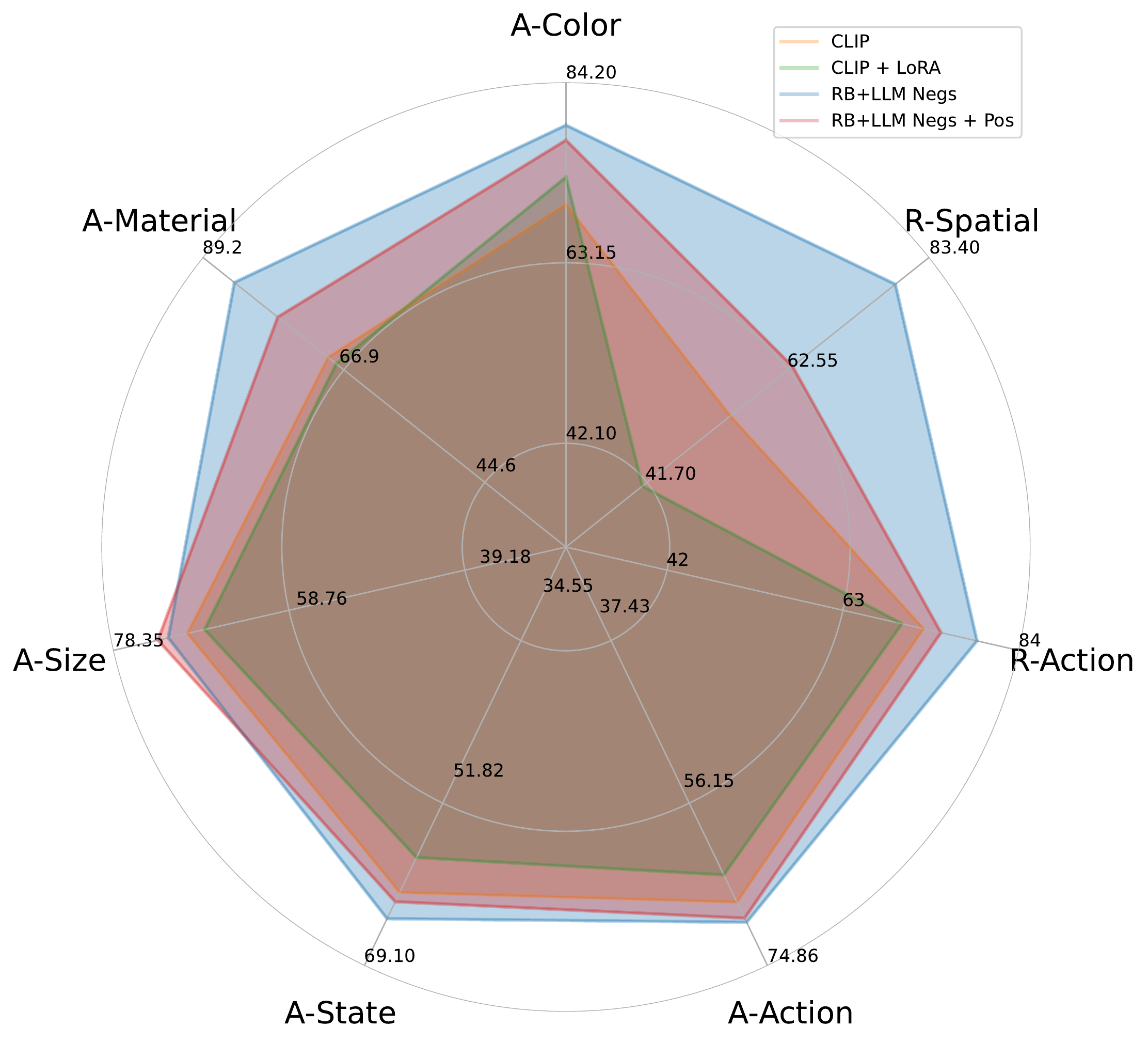}
    \caption{\textbf{CC3M fine-tuning - detailed.} Detailed results of the baselines (CLIP, CLIP+LoRA) and our models (RB+LLM Neg, RB+LLM Negs+Pos) initialized from OpenAI pretrained CLIP ViT-32B and fine-tuned (except CLIP) on CC3M.}
    \label{fig:radar-fine-tune}
    \figvspace
\end{figure}

\subsection{Fine-tuning \vl{} Models}\label{sec:exp-ft-vl-models}
In the following experiments we show the effects of fintuning a pre-trained VL model using our additional data enhancement methods and losses. All experiments are initialized from the official OpenAI CLIP.
We compare our method to two baselines on the VL-Checklist tasks, and on the Zero-Shot image classification task. The first baseline is the original pretrained model without any further training. 
The second, is the same model with the additional LoRA parameters, trained on CC3M (Table~\ref{tab:main_res_vl}a) and LAION (Table~\ref{tab:main_res_vl}g) using the original CLIP loss function. 
Table~\ref{tab:main_res_vl}a shows several configurations of our method compared to the baselines. We see that our method shows significant improvements on all VL-Checklist tasks reaching up to ~15\% improvement. 
Figure~\ref{fig:radar-fine-tune} displays the relative gains on all tasks of the ``Attribute" and ``Relation" tests. It is clear that our gains are across all tests.
These improvements come at a price of some minor degradation to the Zero-Shot performance compared to CLIP. Moreover, when fine-tuning CLIP on the LAION dataset (Table~\ref{tab:main_res_vl}g) we do not see these degradations. 
In \cref{tab:main_res_vl}b we show consistent gains to ones in \cref{tab:main_res_vl}a when finetuning with a stronger (higher ZS) Vit-B/16 CLIP \cite{clip} pre-trained image encoder, observing an even lower drop in ZS performance.
In \cref{tab:main_res_vl}e we show significant gains in fine-tuning CyCLIP on CC3M, comparing fine-tuning before and after integrating our proposed approach, this also comes at no ZS performance cost.

\begin{figure}[b]
    \figvspacetop
    \centering

    \includegraphics[width=0.3\textwidth]{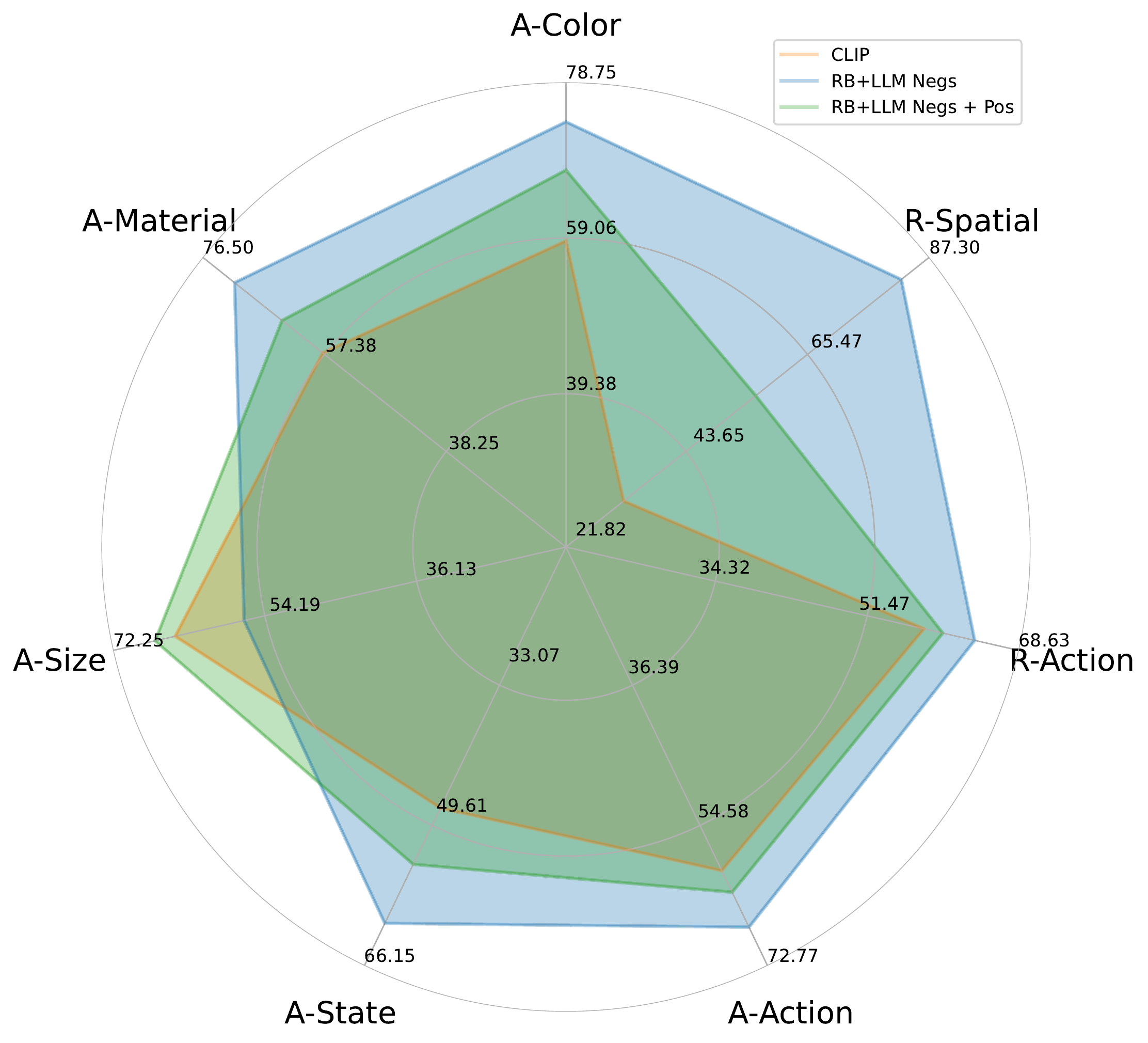}

    \caption{
    \textbf{CC3M train from scratch - detailed.} Detailed results of the baseline CLIP and our models (RB+LLM Neg, RB+LLM Negs+Pos), all CLIP ViT-32B, trained from scratch on CC3M.}
    \label{fig:radar-fs}
    \figvspace
\end{figure}
\subsection{Training from scratch}\label{subsec:fromscratch}
In these experiments, we show the advantage of integrating our approach throughout the whole training procedure. To this end, we train from scratch on CC3M - both CLIP and CyCLIP, either on their own,
or together with our proposed method integrated. Tables - \cref{tab:main_res_vl}c, \cref{tab:main_res_vl}d, and \cref{tab:main_res_vl}f show that, similar to finetining, our method significantly improves both CLIP and CyCLIP \svlc{} capabilities, while still keeping similar ZS performance. Figure~\ref{fig:radar-fs} shows the detailed parsing of the ``Atribute" and ``Relation" tests with consistent gains across the specific tasks. These findings suggest that had we trained our method for many epochs on a large dataset, such as the full LAION-9B, we would reach similar ZS performance while greatly enhancing the \svlc{} performance of the model.

\secvspace
\section{Ablations}\label{sec:ablations}
\secvspace
In this section we will examine several aspects and components of our proposed methods.
\secvspace
\subsection{Negative Losses}
\secvspace
Section~\ref{sec:losses} details our additional loss functions which utilize our generated texts. Specifically, \cref{eq:neg} describes the loss which contrasts a given example \textbf{only} with its generated negatives. Table~\ref{tab:ablation-neg} clearly shows the importance of this loss as opposed to simply adding the negative examples to the original contrastive loss (\cref{eq:contrastive}). Without explicitly forcing the network to attend to the small changes in the text (\cref{eq:neg}), the generated negative examples, when inserted only to the contrastive loss (\cref{eq:contrastive}) do not provide the desired gains over the baseline.

\secvspace
\subsection{Component analysis}
\secvspace
In Section~\ref{sec:method} we describe several components of our proposed method. Specifically we present our RB negative generation (Sec.~\ref{sec:rb_neg}), our LLM-based negative generation (Sec.~\ref{sec:auto_neg}), and our LLM-based analogy generation, refered to here as ``Pos" (Sec.~\ref{sec:pos}). 
\cref{tab:ablation_comp_att_rel} provides a detailed analysis and comparison of the contribution of each component to the final result. Through this analysis we see two contradicting forces between the \svlc{} capabilities and the original Zero-Shot performance. We can see that each negative generation method plays a crucial role in some of the tasks while the use of both is usually the best performing option. On the other hand, the LLM-based analogy generation stabilizes the Zero-Shot performance and mitigates the drop with respect to the baseline. The joint version of all components (``Ours Combined") exhibits a good trade-off between the two contradicting forces.


\secvspace
\section{Conclusions}
\secvspace
We have presented a data-driven technique for enhancing the performance of \vl{} models in the important task of \svlc{} understanding without sacrificing their impressive ZS object recognition capabilities. Our proposed method attains significant gains in multiple experiments on a variety of base \vl{} models and datasets. It builds upon the modeling strength and knowledge of language structure to teach this structure to \vl{} models in an orthogonal
way, suggesting wide applicability to existing or future \vl{} models.

While attaining impressive gains in \svlc{} understanding, we believe the small drop observed in ZS performance could be further reduced.
An additional possible extension of our work is using more sophisticated sequence generation techniques for improving our batch data. One may combine annotation efforts with a language model to get improved data for training and evaluation \cite{liu2022wanli}. Another possibility is adding a corrector \cite{Welleck2022Generating} in the training that validates whether the \vl{} model learns the correct concepts or not. We leave all of these exciting directions to future work.

\section*{Acknowledgements}
\footnotesize
We would like to thank Donghyun Kim for his invaluable help in this work.
This material is based upon work supported by the Defense Advanced Research Projects Agency (DARPA) under Contract No. FA8750-19-C-1001. Any opinions, findings and conclusions or recommendations expressed in this material are those of the author(s) and do not necessarily reflect the views of the Defense Advanced Research Projects Agency (DARPA).
This material is based upon work supported by the ‘Data Science grant from the Israeli Council of Higher Education’.
This research or RG was supported by ERC-StG SPADE grant no. 75749.

\clearpage
{\small
\bibliographystyle{ieee_fullname}
\bibliography{egbib}
}

\end{document}


\title{Supplementary Material for Teaching Structured Vision \& Language Concepts to Vision \& Language Models}

\newcommand{\minisection}[1]{\noindent{\textbf{#1}.}}
\author{
    Sivan Doveh$^{1,2}$,
    Assaf Arbelle$^{1}$,
    Sivan Harary$^{1}$,
    Eli Schwartz$^{1,3}$,
    Roei Herzig$^{1,3}$,\\
    Raja Giryes$^{3}$,
    Rogerio Feris$^{4}$,
    Rameswar Panda$^{4}$,
    Shimon Ullman*$^{2}$,
    Leonid Karlinsky\thanks{Equal contribution}\hspace{-1pt}*$^{4}$\\
    \\
    \tt\small
    $^{1}$IBM Research,
    $^{2}$Weizmann Institute of Science,
    $^{3}$Tel-Aviv University, 
    \tt\small
    $^{4}$MIT-IBM Watson AI Lab
}
\maketitle

\section{Rule-Based Negative Generation}\label{supsec:rb}
In this section we describe in detail the process of the Rule-Based negative generation discussed in Section~3.1.1 of the main paper. 
The Rule-Based method is a simple yet effective method for generating negative examples of a specific \svlc{} type. We do this by first collecting a list of all words related to the desired \svlc{} type, this can be done by a simple internet search. 
For every sentence in our dataset we can compare the words in the sentence with the list. If the sentence contains a word from the list we randomly replace it with a different word from the same list. 
\begin{algorithm}
\caption{A pseudo-code for generating Rule-Based negative examples}
\begin{algorithmic}[1]
\STATE Let $\mathcal{L}$ be a list of words
\STATE Let $\mathcal{T}$ a dataset of sentences
\FORALL{$t\in \mathcal{T}$} 
\STATE Let $W$ be all words in $t$
\FORALL{$w\in t$}
\IF{$w\in \mathcal{L}$}
\STATE Sample $w'$ from $\mathcal{L}$
\STATE Replace $w$ with $w'$ in $t$
\STATE break~~~~\# we replace only a single word
\ENDIF
\ENDFOR
\ENDFOR
\end{algorithmic}
\label{alg:rb}
\end{algorithm}

For example, when creating the rule-based negatives for \textbf{colors} we collected the list: teal, brown, green, black, silver, white, yellow, purple, gray, blue, orange, red, blond, concrete, cream, beige, tan, pink, maroon,
olive, violet, charcoal, bronze, gold, navy, coral, burgundy, mauve, peach, rust, cyan, clay, ruby, and amber. 
Then applying the RB-negatives algorithm on a given sentence ``A \textbf{blue} car on the road" could randomly change the color \textbf{blue} to \textbf{beige} to get: ``A \textbf{beige} car on the road". The pseudo-code for RB-negatives is described in Algorithm~\ref{alg:rb}.
Similarly, we have RB-negatives generation set-up for several \svlc{}s, namely: colors, materials, states, sizes, and actions.


\section{LLM-Based Negative generation}\label{supsec:llm-neg}
\begin{algorithm}
\caption{The pseudo-code for generating LLM-based negative examples}
\begin{algorithmic}[1]
\STATE Let $\mathcal{T}$ be a dataset of sentences
\FORALL{$t\in \mathcal{T}$} 
\STATE Parse $t$ using spacy
\STATE Randomly choose a part of the sentence (POSTAG)
\STATE Randomly choose word $w \in t$ that has the chosen POSTAG
\STATE Replace $w$ with $<$MASK$>$ token in $t$
\STATE $W'\gets$ \text{unmask using DistilRoBERTa} {}
\STATE Remove $w$ from $W'$
\STATE Randomly select $w'$ from $W'$
\STATE Replace $w$ with $w'$ in $t$
\ENDFOR
\end{algorithmic}
\label{alg:llm-neg}
\end{algorithm}
As opposed to the RB negative generation (SupSec~\ref{supsec:rb}), the LLM-based negative generation does not require a human definition of the set of valid negatives.
This method, introduced in Section~3.1.2 of the main paper, can be broken into three steps using two components of language modeling.
First, we extract the linguistic parts of the sentence such as nouns, adjectives, verbs, etc. For this step, we used the spacy~\cite{spacy} "en-core-web-sm" python package. 
Then we randomly select which part to change and randomly choose a word of that category. 
We replace the selected word using Distil-Roberta~\cite{roberta,distillbert,distillroberta} mask-filling capabilities. We replace the selected word with the masking token $<$MASK$>$ and input the new sentence to the Distil-Roberta model which outputs several candidates for valid words. We select one word from the list, filtering out the original word. 
For example, the sentence: ``Two kids playing in the park" was parsed and the verb ``playing" was randomly selected. 
We mask out this word to get: ``Two kids $<$MASK$>$ in the park". 
The model then predicts several valid words: sitting, playing, eating, drawing, running. We randomly select one of these possibilities while excluding the original word ``playing" to get the negative caption: ``Two kids eating in the park". The pseudo-code for LLM-based negatives generation is described in Algorithm~\ref{alg:llm-neg}.





\section{Text Analogies via LLM Prompting}\label{supsec:pos}

This method, as discussed in Section~3.1.3 of the main paper, aims to generate a semantically similar sentence with different wording than the original sentence. When generating negative examples (Sections~\ref{supsec:rb},\ref{supsec:llm-neg}) the goal was to make \textbf{minimal} changes in the original sentence while significantly changing the meaning. However, in this case we require the exact opposite, i.e. major changes in the sentence while still keeping the same semantic meaning. We generate these sentences using BLOOM~\cite{bloom}. We prompt the model with the following caption: \textit{``a woman standing left to a sitting cat is semantic similar to a cat standing right to a woman. a baby crying to the right of a box is semantic similar to a box placed to the left of a crying baby. a man sitting to the right of a dog is semantic similar to a dog sitting to the left of a man. a blue boat is semantic similar to a boat that is blue."} Followed by:  \textit{\textbf{CAPTION} is semantic similar to"}, where \textit{\textbf{CAPTION}} is the sentence we want to find analogy for. The output of the BLOOM model is a continuation of the sentence in the spirit of the initial prompt. 


\section{Analysis and Ablation Study}

\subsection{Individual Dataset Analysis}
As discussed in the Dataset section (Section~4.1) of the main paper, the VL-Checklist~\cite{vlc} evaluation dataset is comprised of four different sets, Visual Genome (VG)~\cite{vg}, VAW~\cite{vaw}, SWIG~\cite{swig}, and HAKE~\cite{hake}. 
In the main paper, we presented the results over a combination of all the datasets evaluated jointly. As promised in the main paper, a more detailed analysis partitioned according to the individual datasets comprising VL-Checklist is available in tables~\ref{FT_VG_Obj}-\ref{FS_VAW_Att}.
Tables~\ref{FT_VG_Obj}-\ref{FT-Vaw} show the results of the CC3M fine-tuning experiments, corresponding to Table~1a in the main paper. 
Tables~\ref{FS_VG_obj}-\ref{FS_VAW_Att} show the results of the models trained from scratch on CC3M, corresponding to Table~1c in the main paper. 
Note that not all aspects of the tests are available for all datasets, for example, VAW evaluation only includes the ``Attribute" aspects, i.e. color, size, material, state, and action. For each dataset we include evaluations on all of its available aspects.

\subsection{Error Analysis}
\begin{figure*}
   
\begin{tabular}{cc}
     
      \rotatebox{90}{~~~~~~~~~~~~~~~~A-Action}&\includegraphics[width=0.3\textwidth]{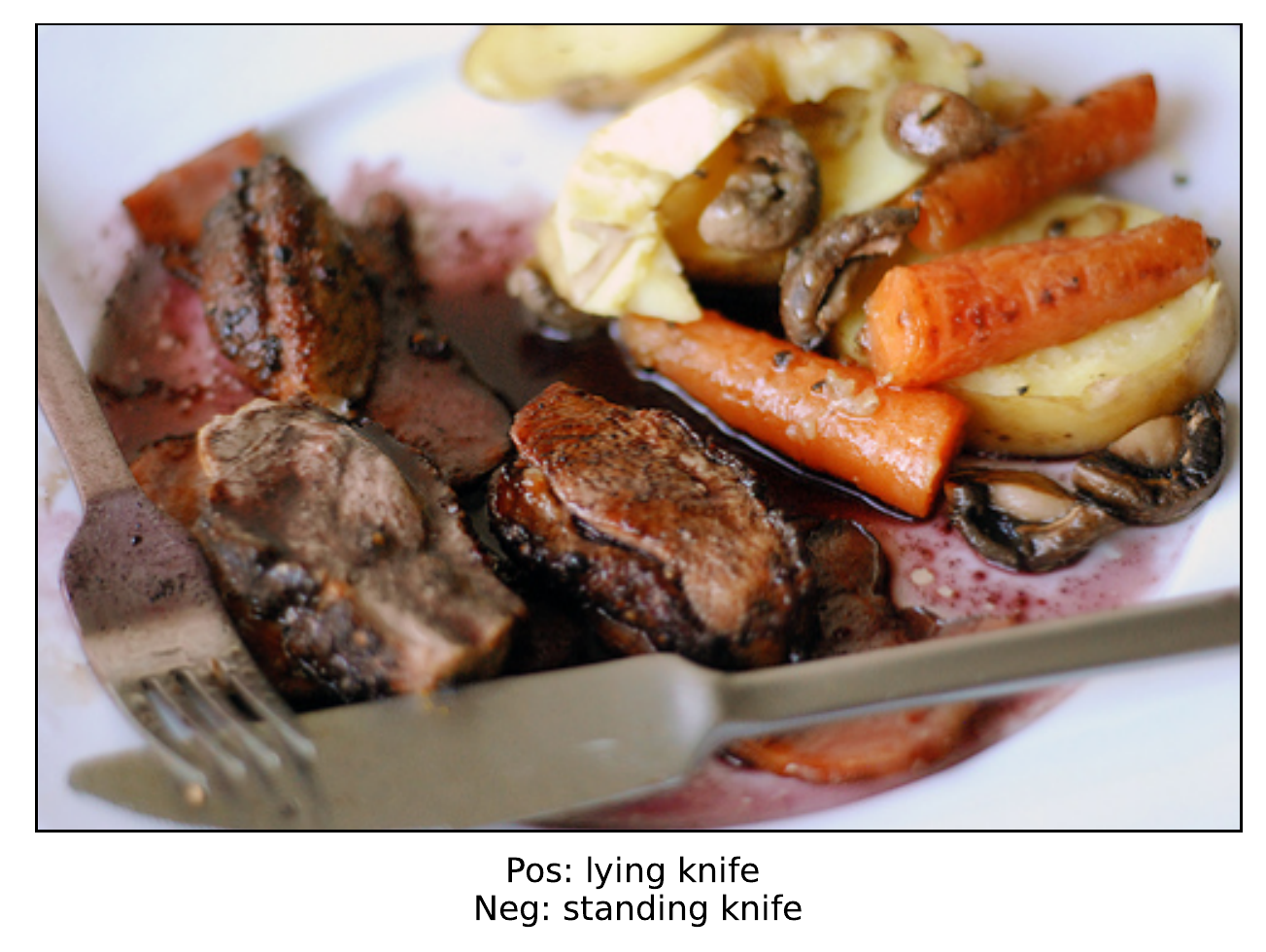}\includegraphics[width=0.3\textwidth]{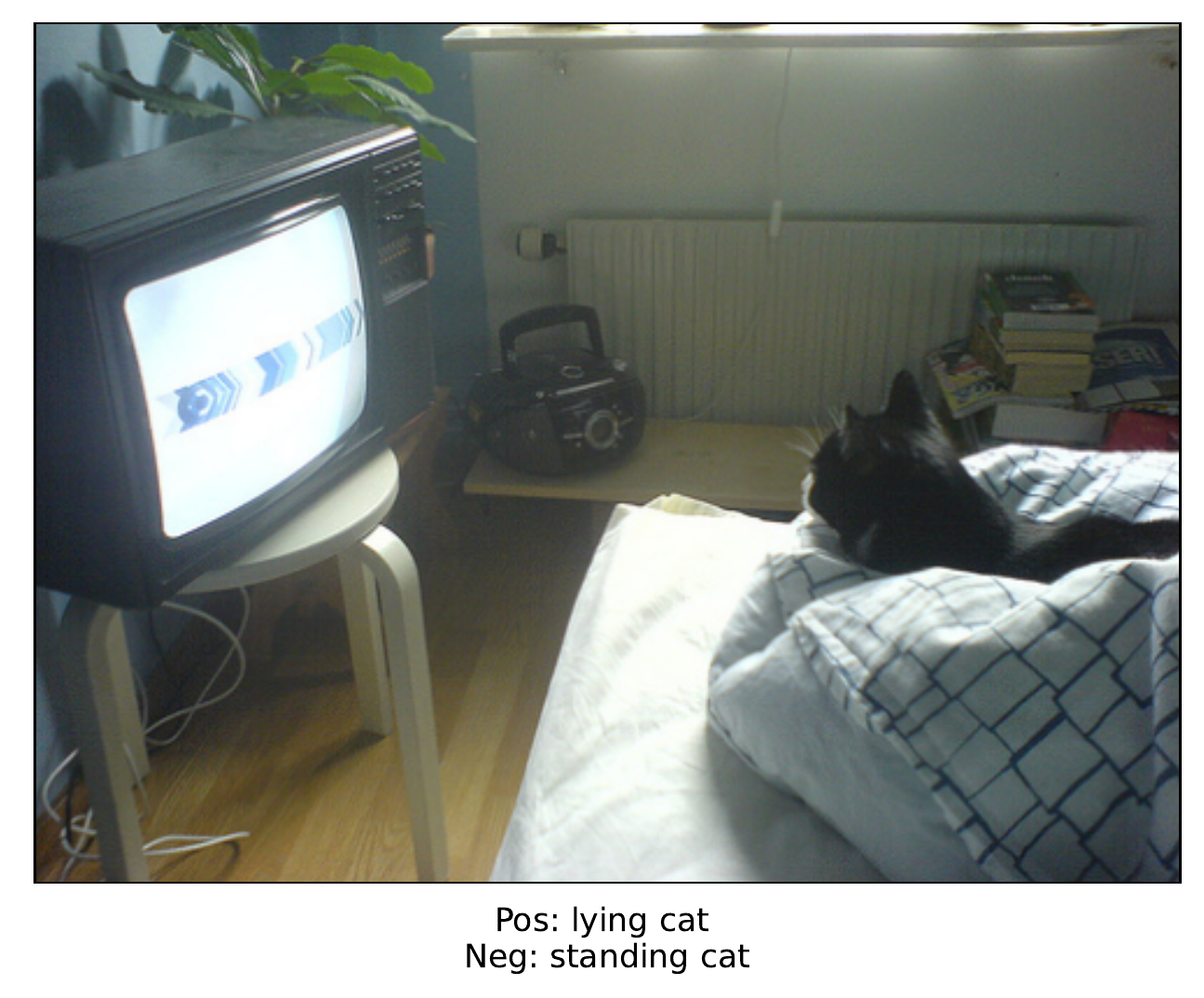}\includegraphics[width=0.3\textwidth]{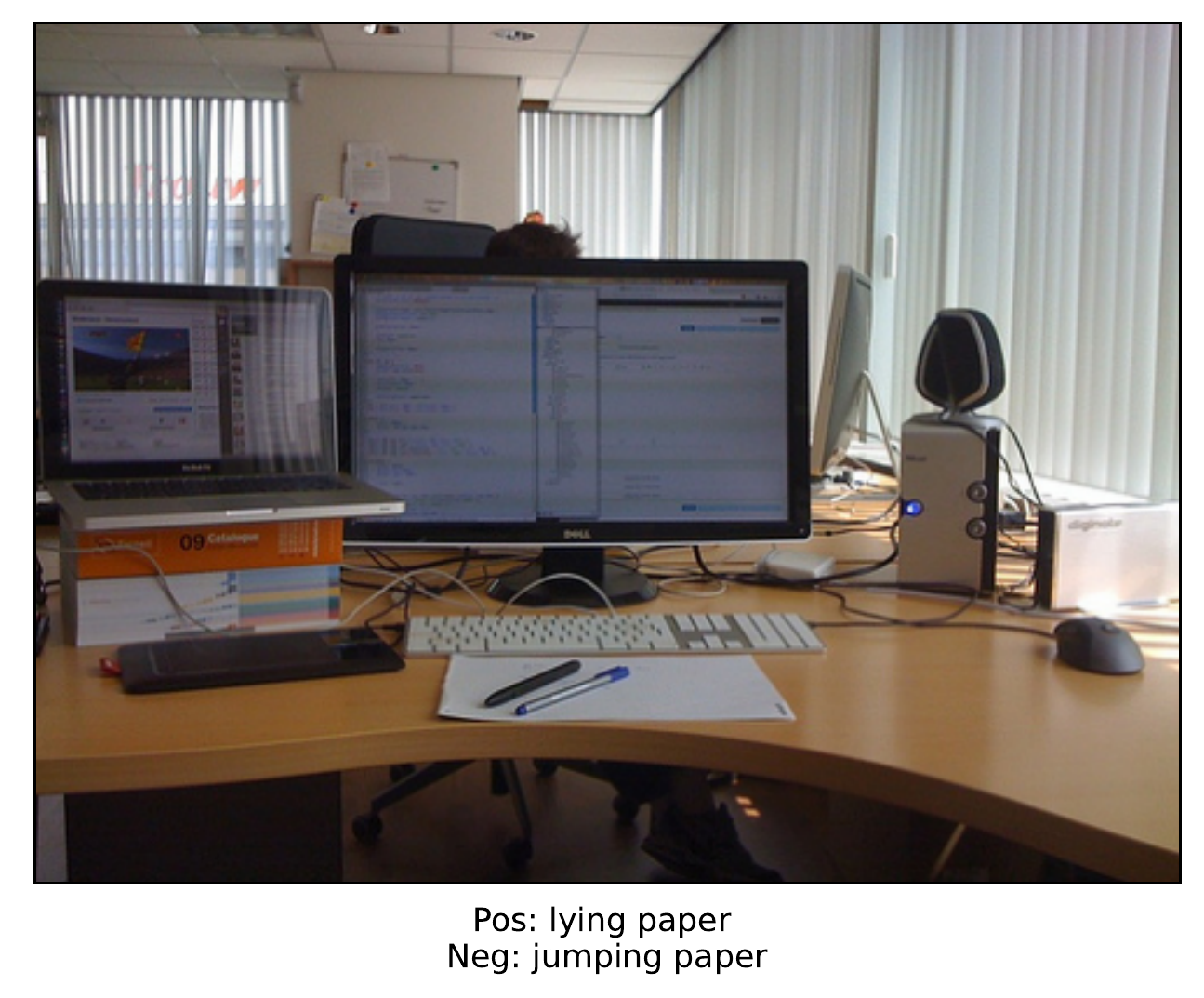}\\
     
     \rotatebox{90}{~~~~~~~~~~~~~~~~~~~A-Color}&\includegraphics[width=0.3\textwidth]{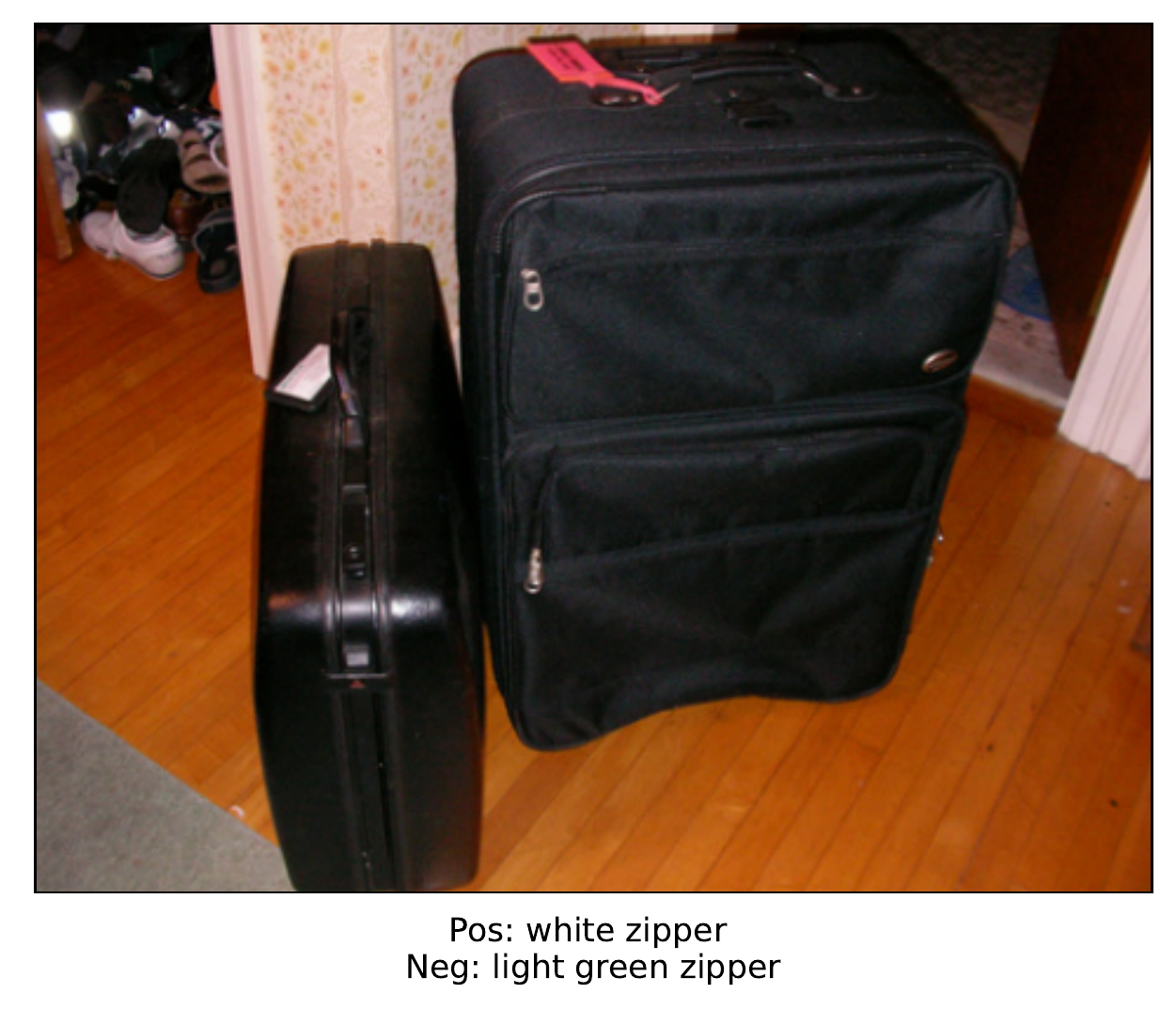}\includegraphics[width=0.3\textwidth]{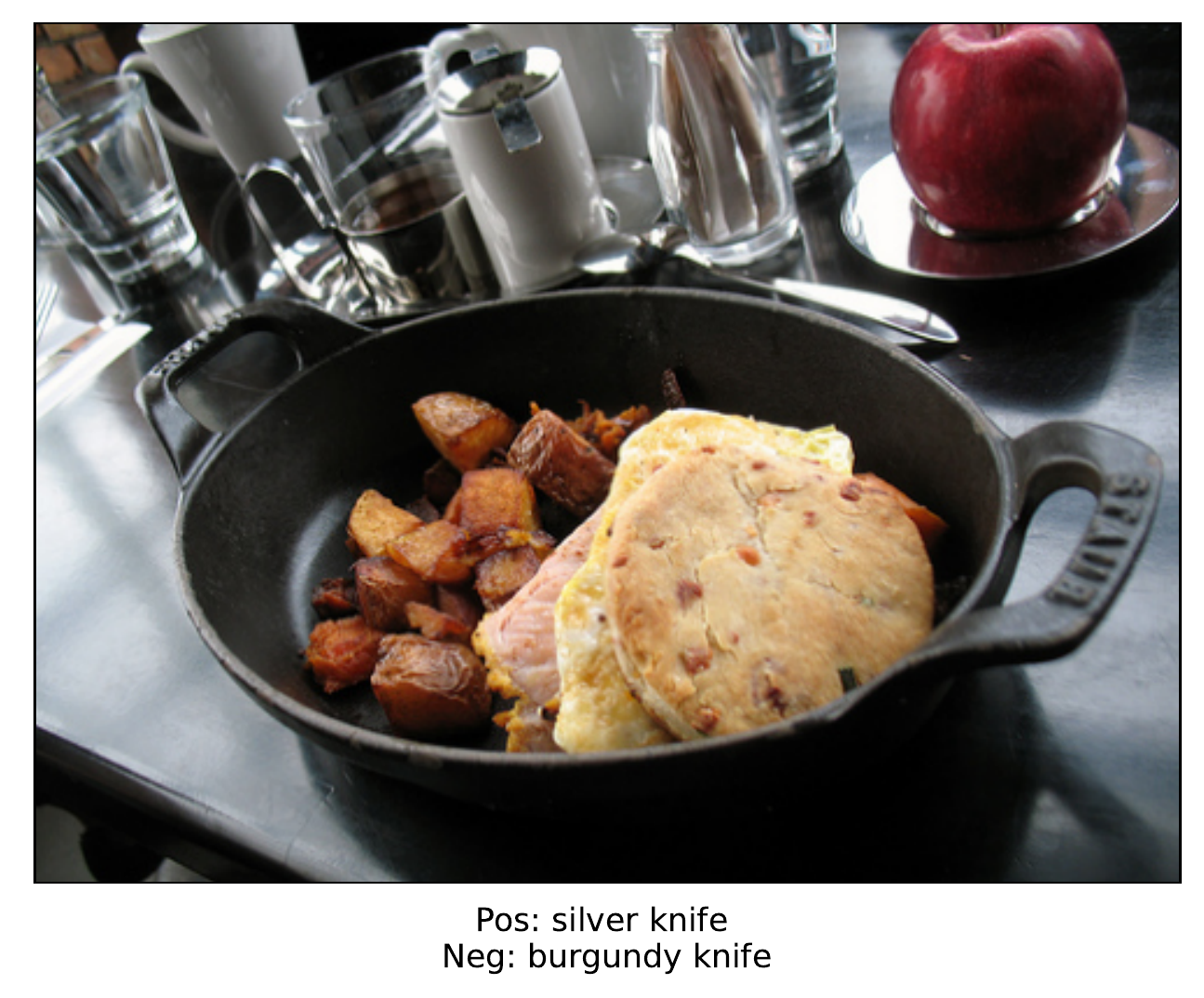}\includegraphics[width=0.3\textwidth]{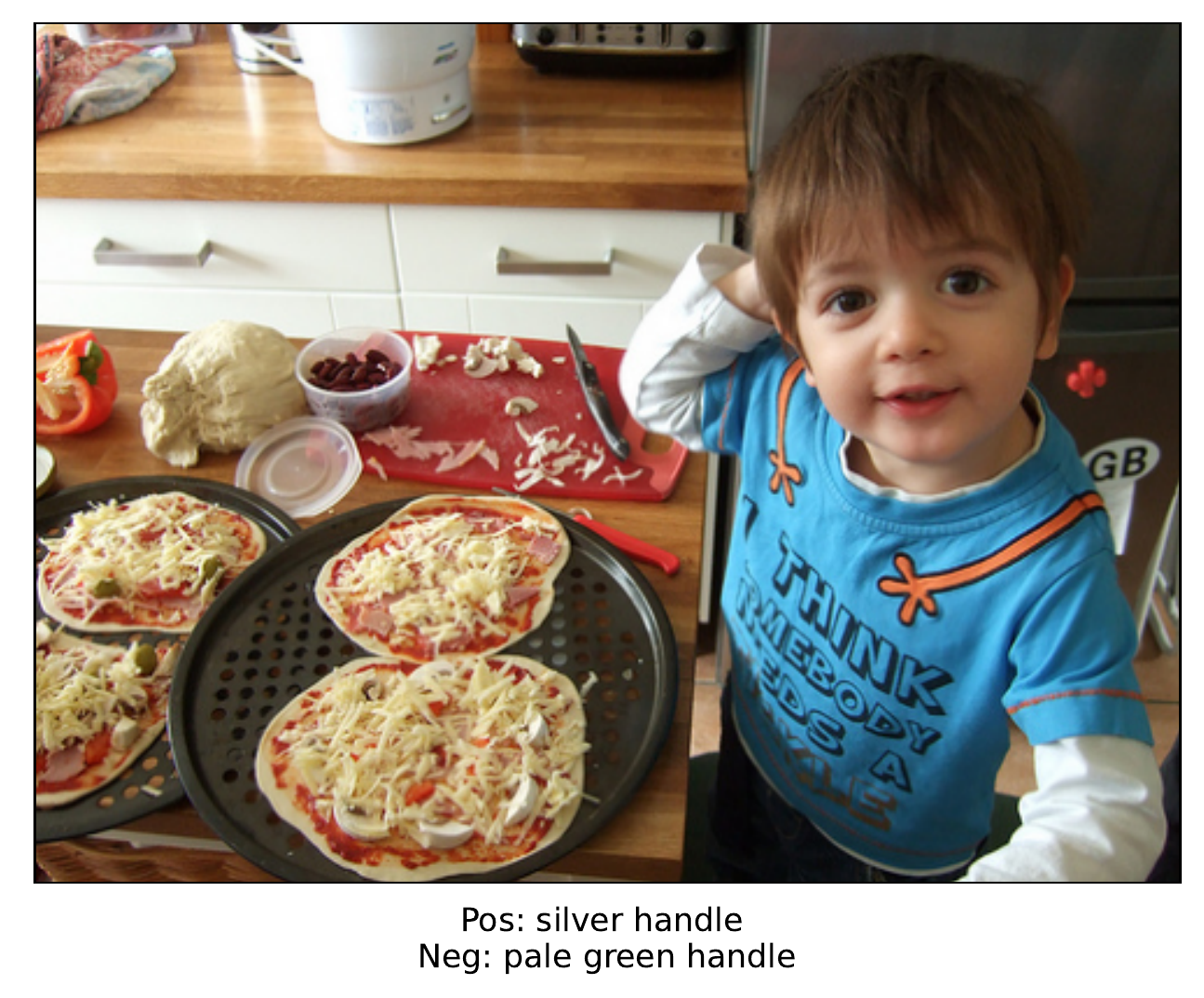}\\

    \rotatebox{90}{~~~~~~~~~~~~~~~~~~~A-Material}&\includegraphics[width=0.3\textwidth]{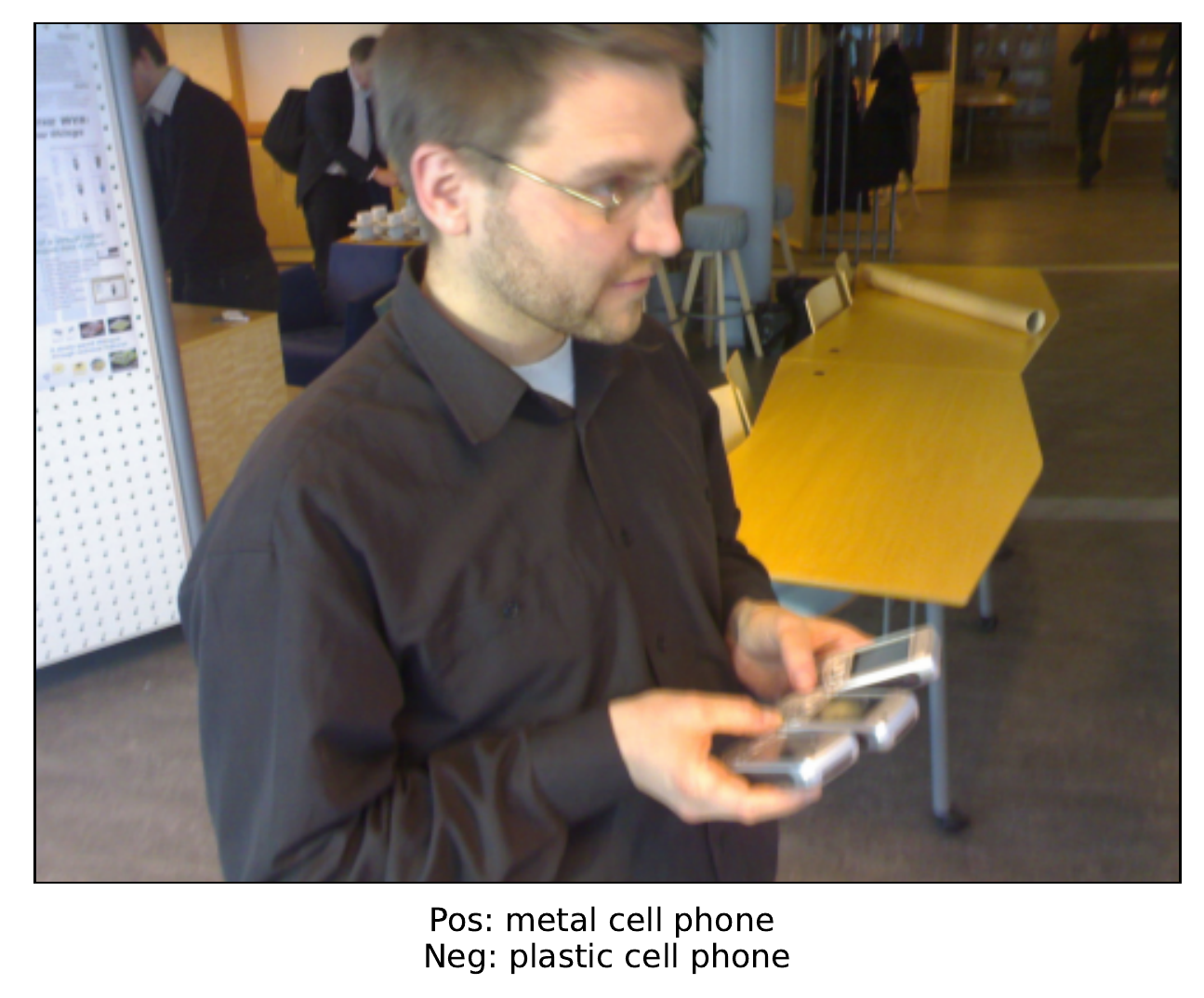}\includegraphics[width=0.3\textwidth]{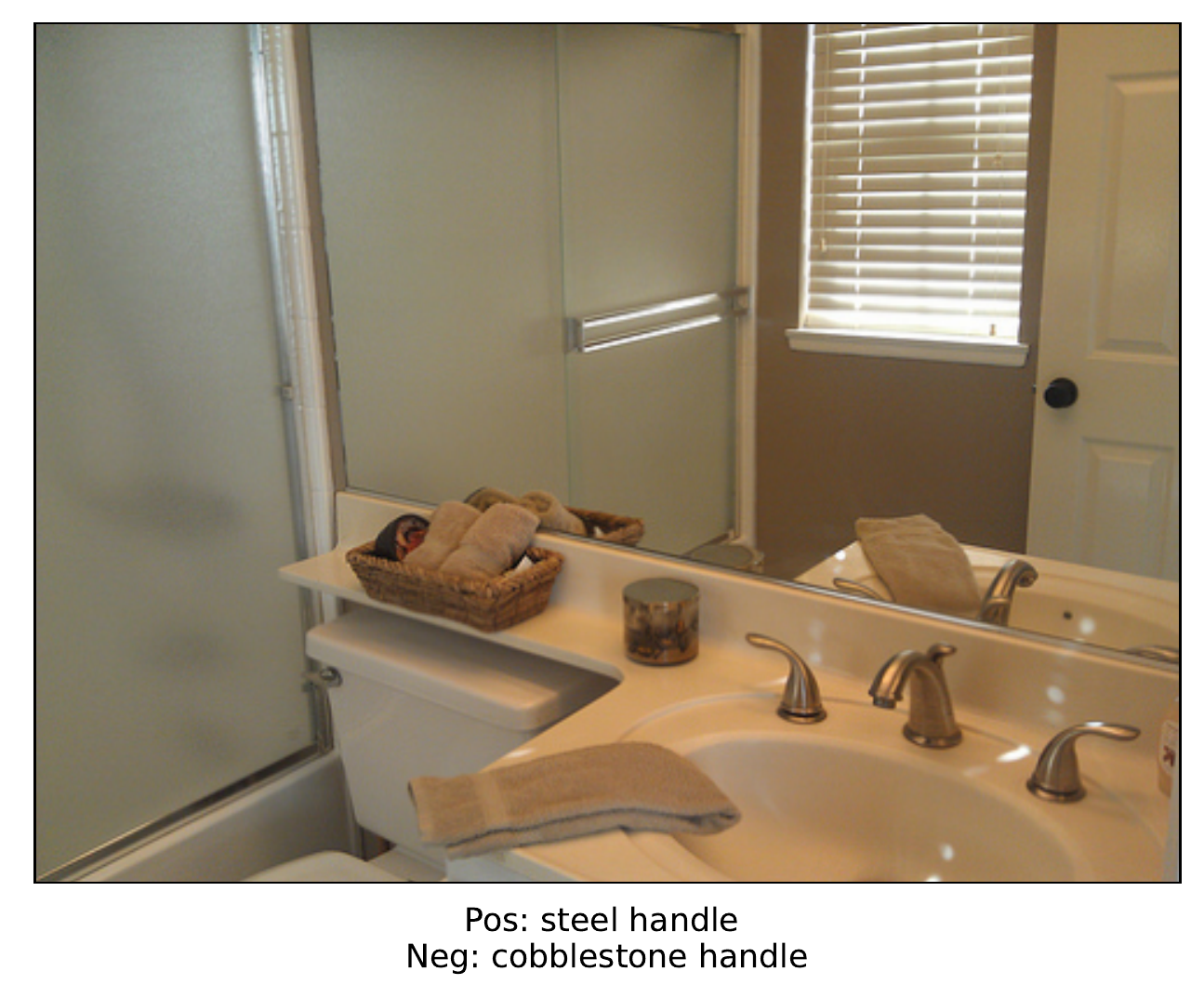}\includegraphics[width=0.3\textwidth]{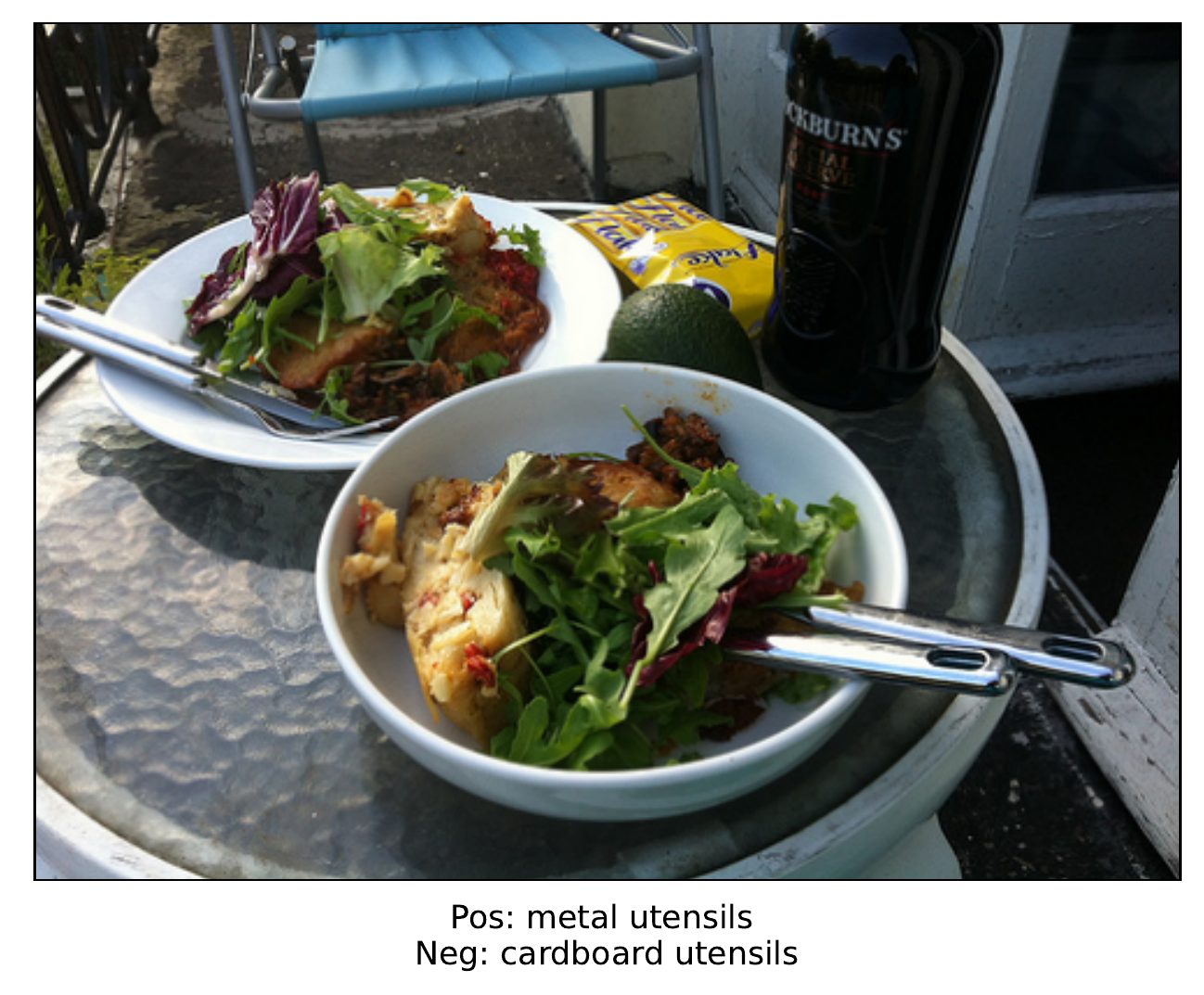}\\

    \rotatebox{90}{~~~~~~~~~~~~~~~~~~~~~~~~~~~~~~~~~~~~~~~~~~~~~A-State}&\includegraphics[width=0.3\textwidth]{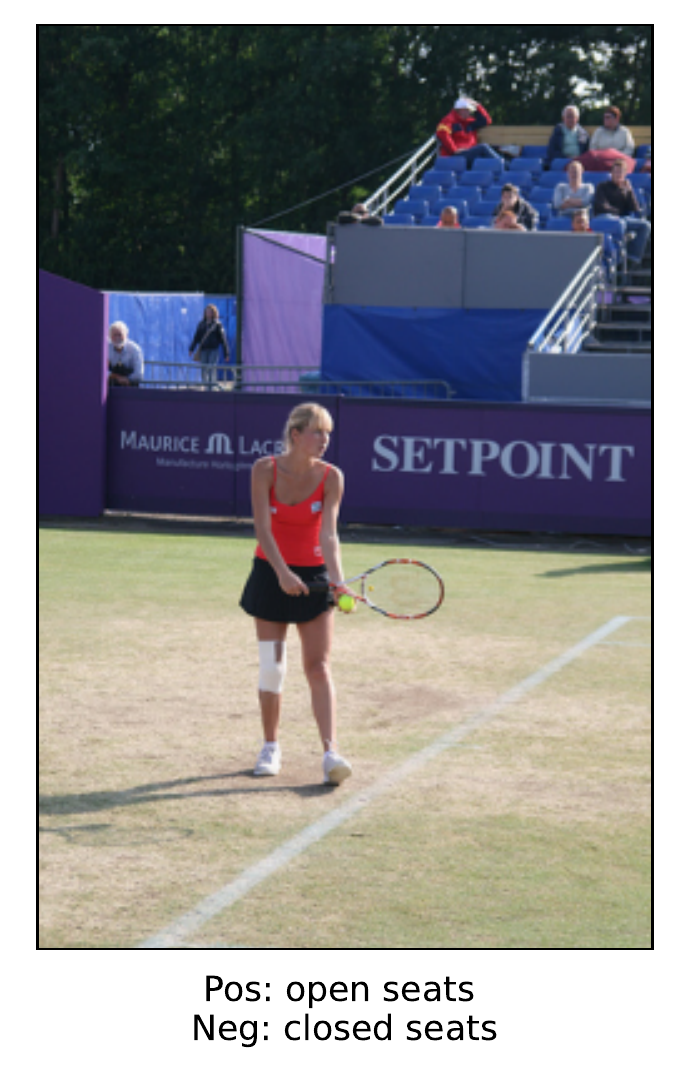}\includegraphics[width=0.3\textwidth]{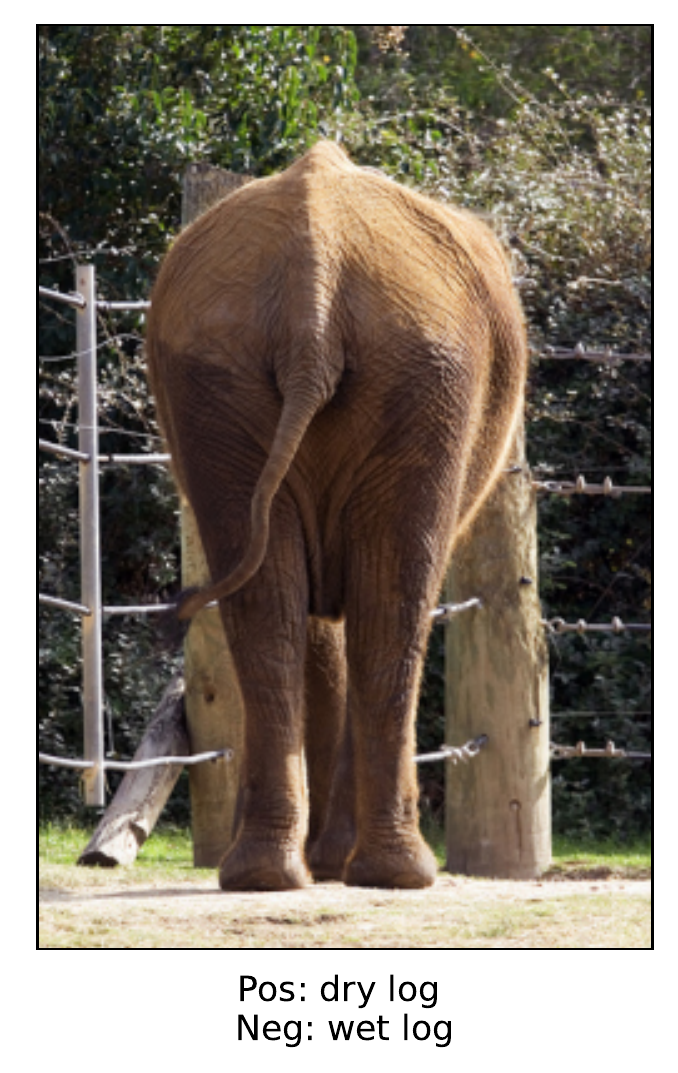}\includegraphics[width=0.3\textwidth]{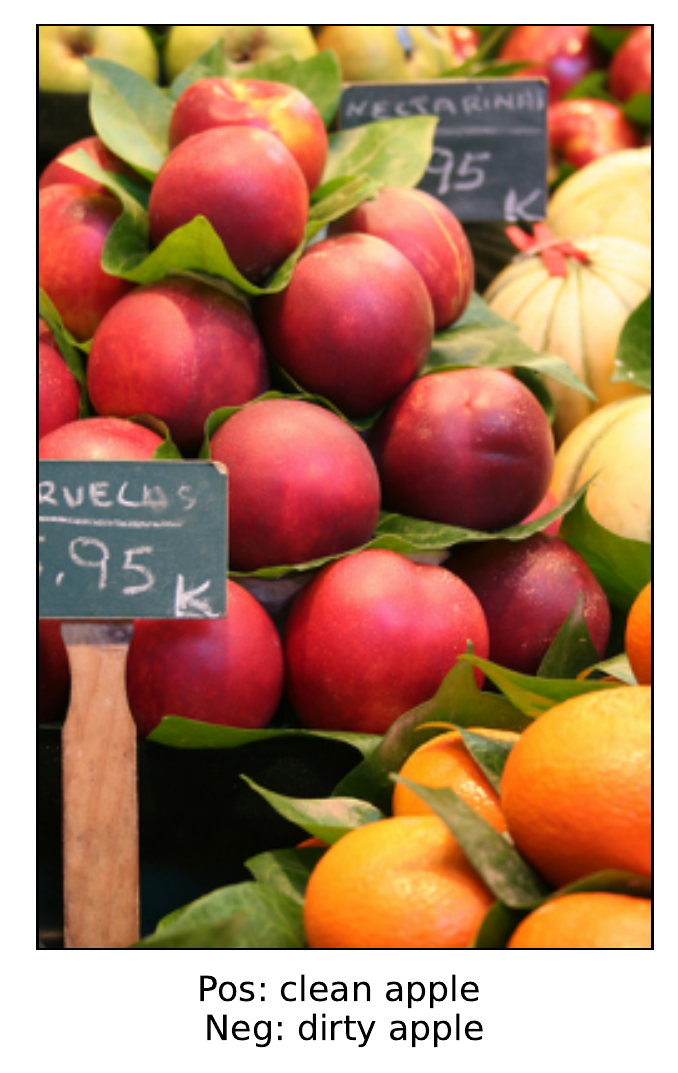}\\

 \end{tabular}
 \caption{Examples where our model correctly chooses the positive caption, while the CLIP baseline fails and incorrectly chooses the negative caption. We show the respective positive and negative captions underneath each image.}
 \label{fig:tf-1}
 \end{figure*}

\begin{figure*}
   
\begin{tabular}{cc}

    \rotatebox{90}{~~~~~~~~~~~~~~~~A-Size}&\includegraphics[width=0.3\textwidth]{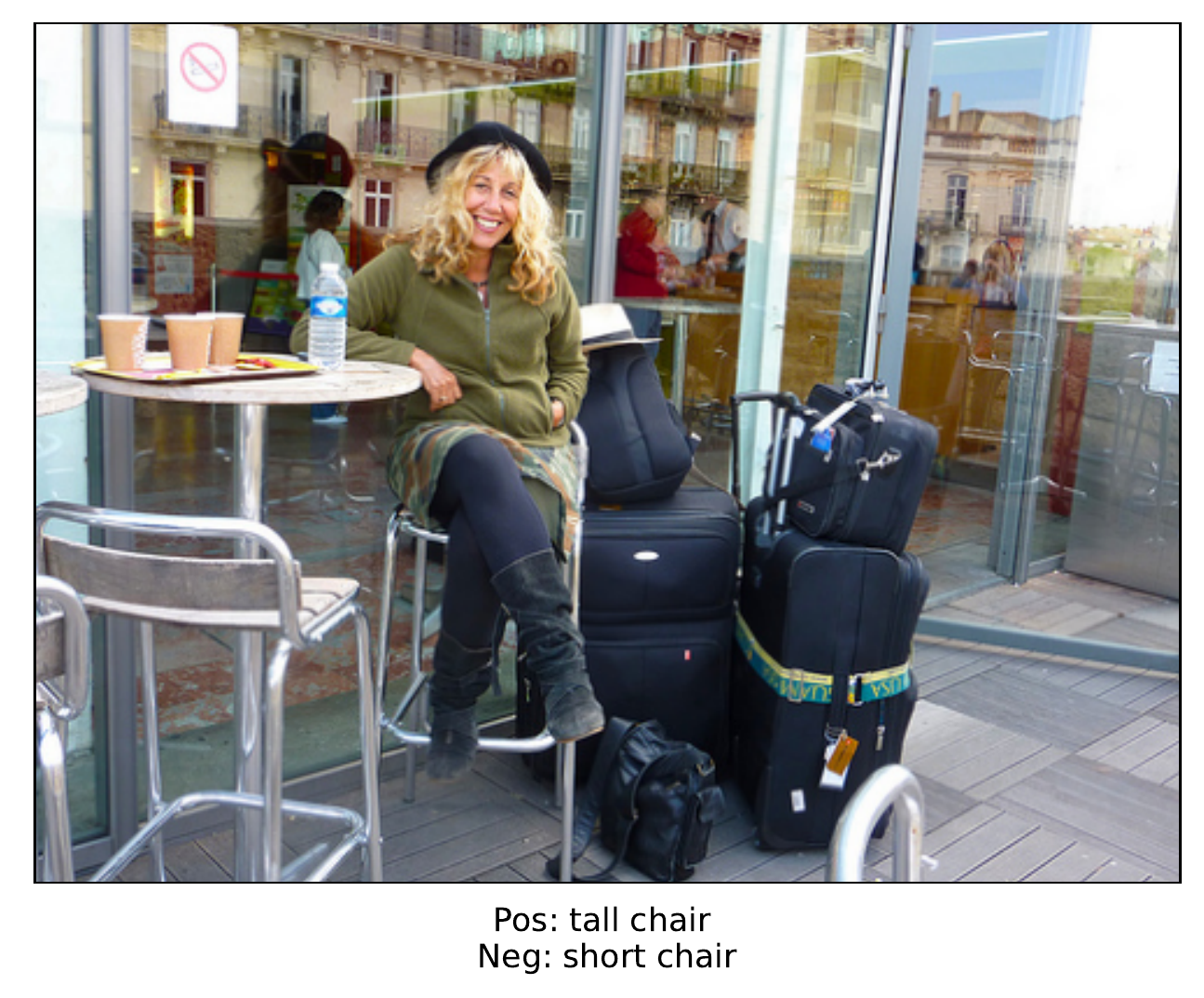}\includegraphics[width=0.3\textwidth]{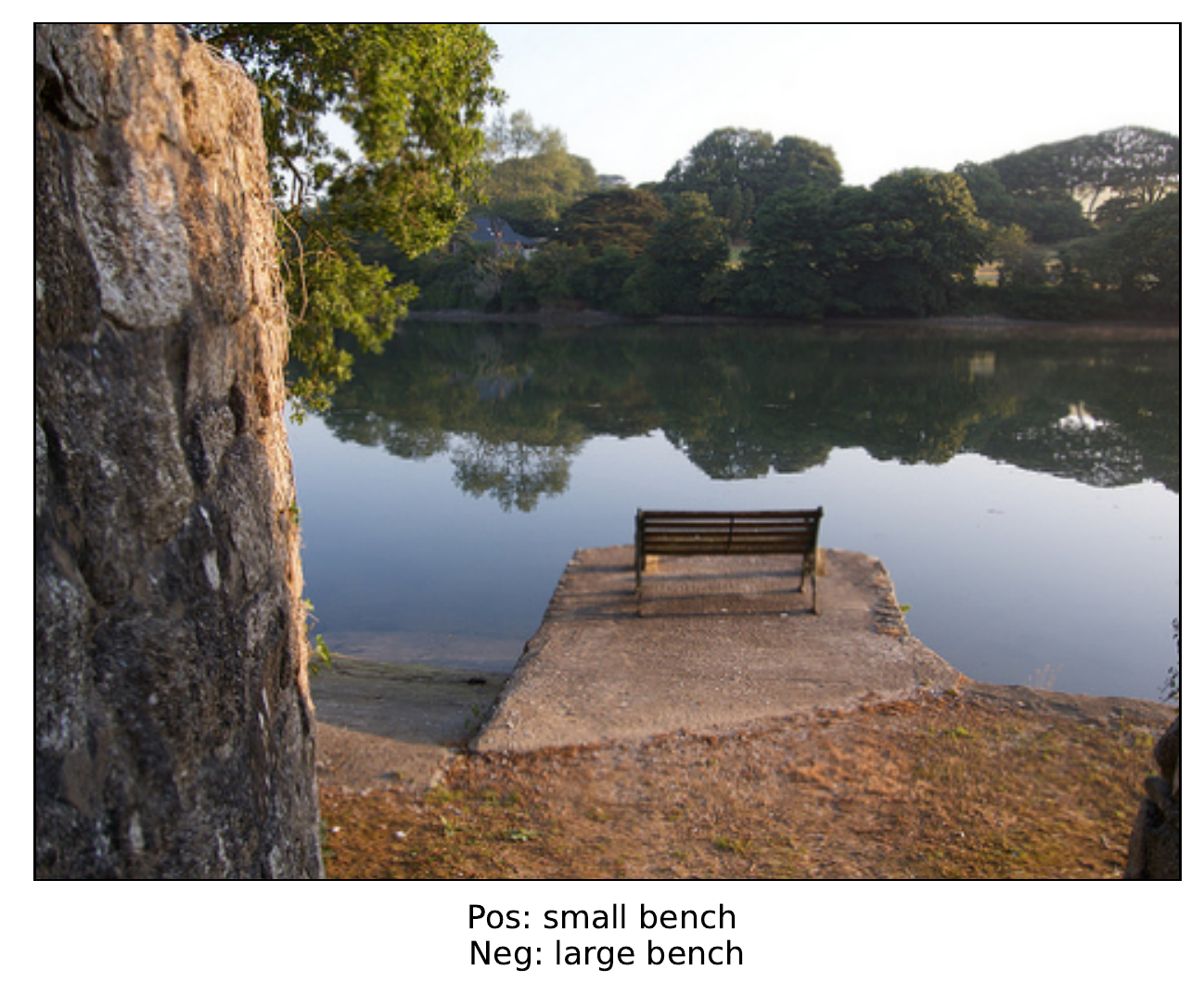}\includegraphics[width=0.3\textwidth]{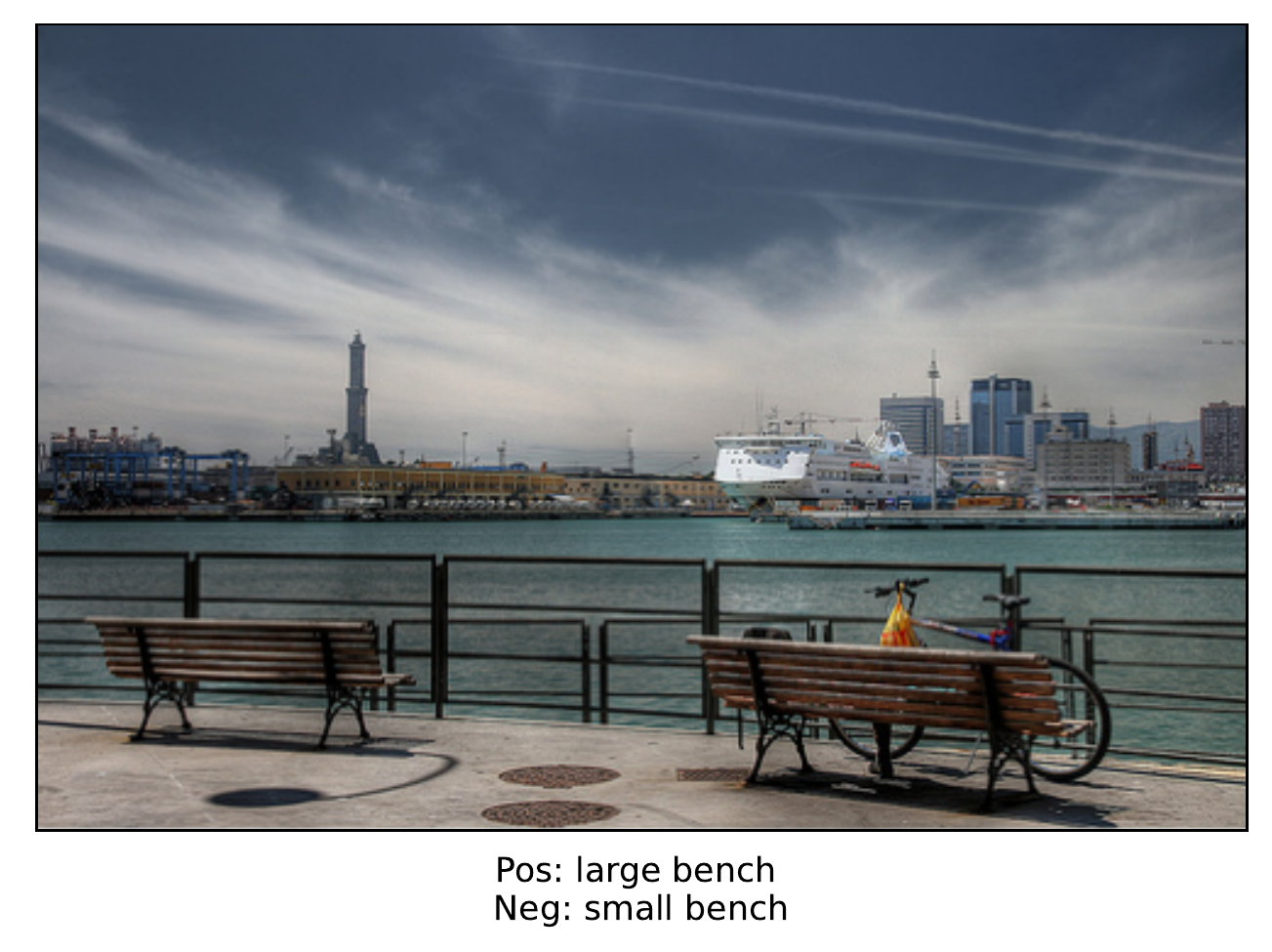}\\
    
    \rotatebox{90}{~~~~~~~~~~~~~~~~R-Action}&\includegraphics[width=0.3\textwidth]{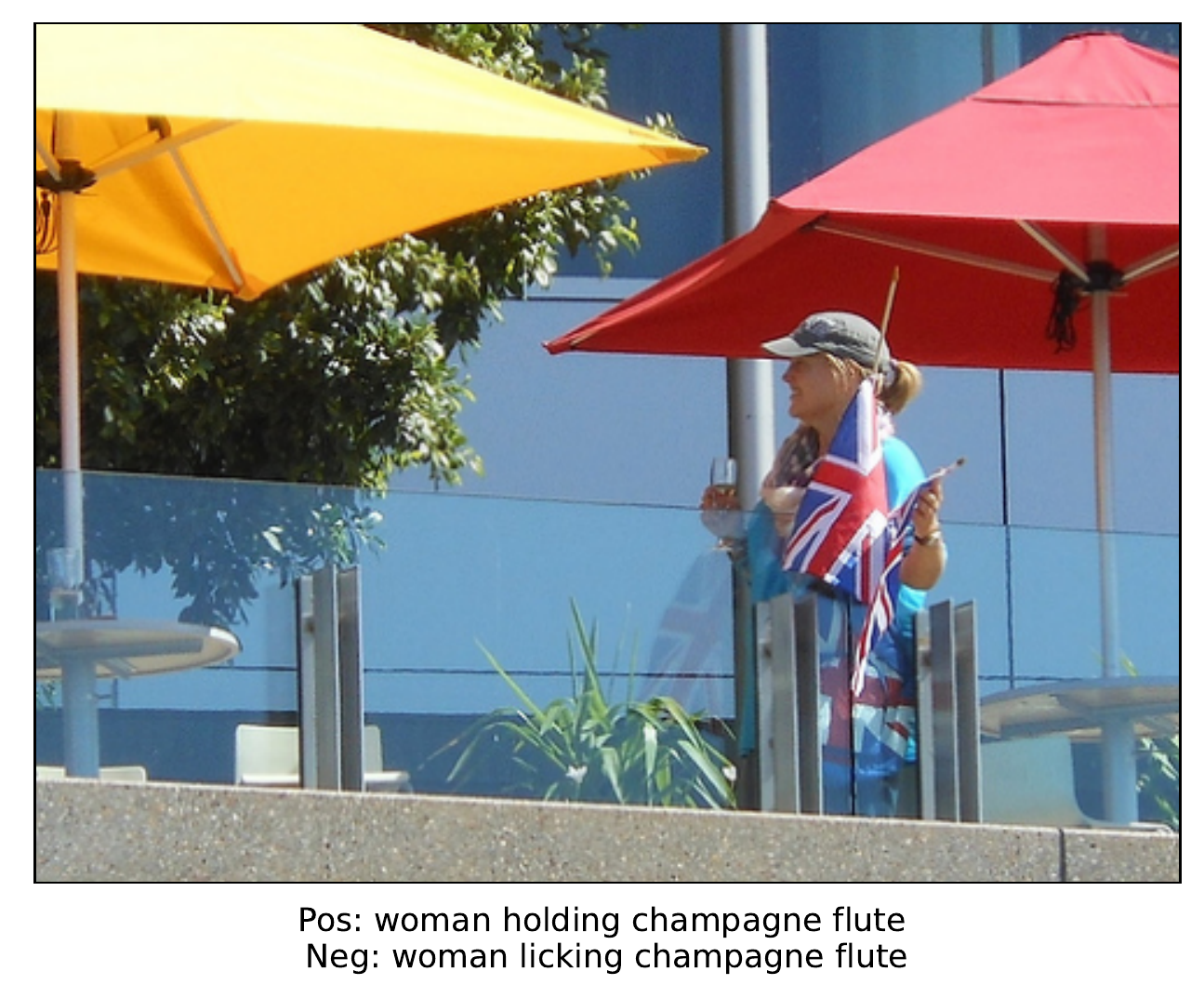}\includegraphics[width=0.3\textwidth]{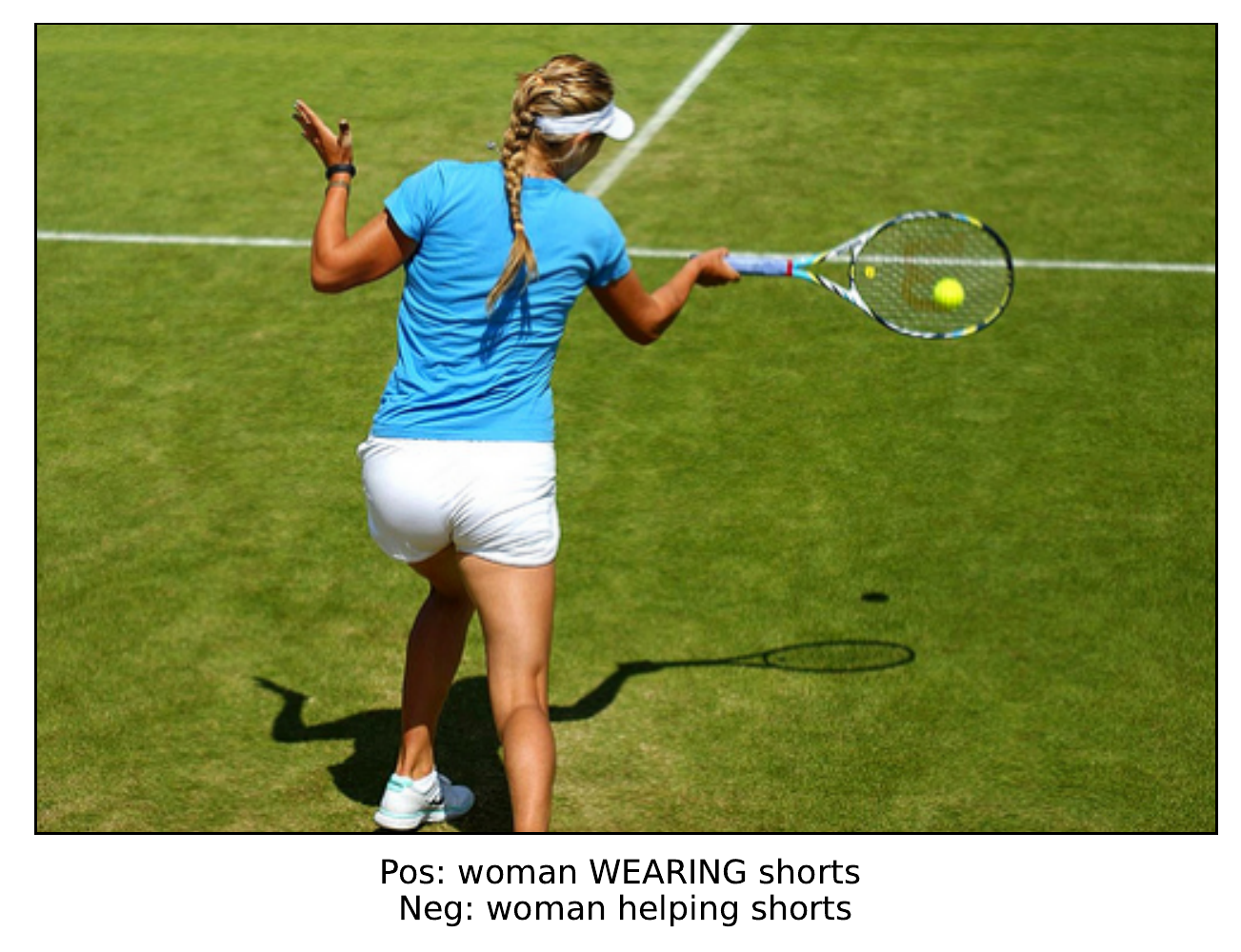}\includegraphics[width=0.3\textwidth]{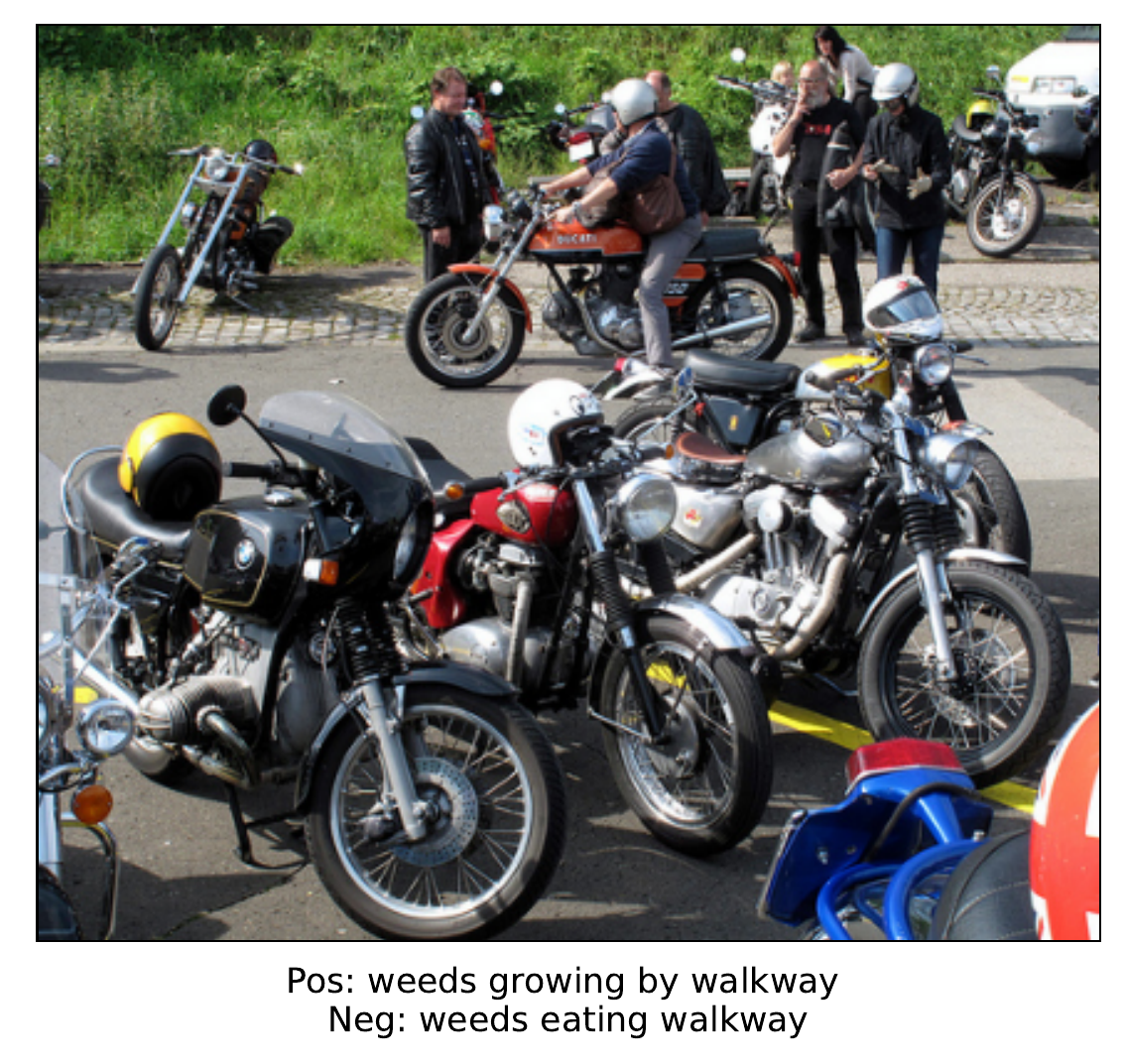}\\
     
    \rotatebox{90}{~~~~~~~~~~~~~~~~R-Spatial}&\includegraphics[width=0.3\textwidth]{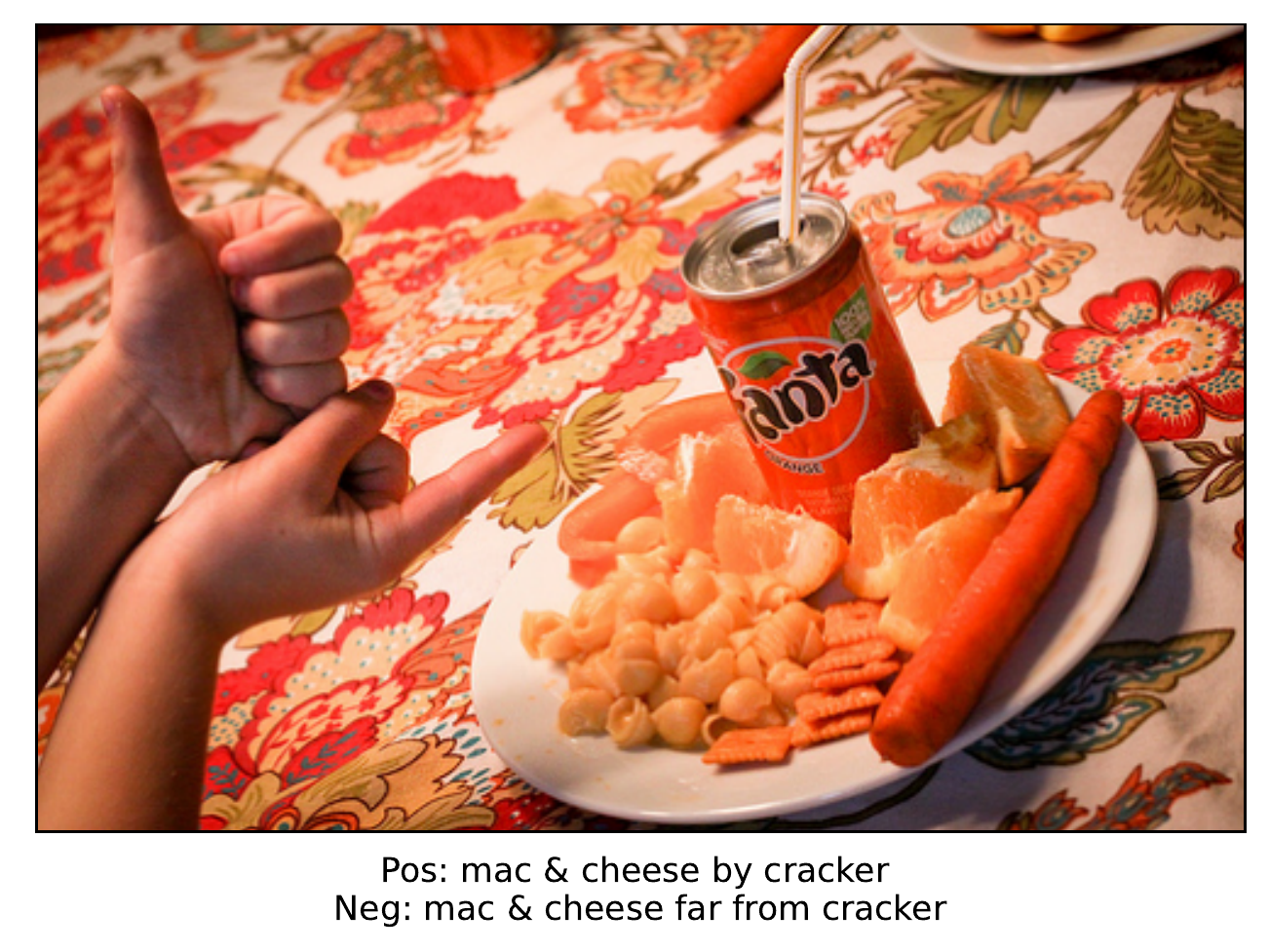}\includegraphics[width=0.3\textwidth]{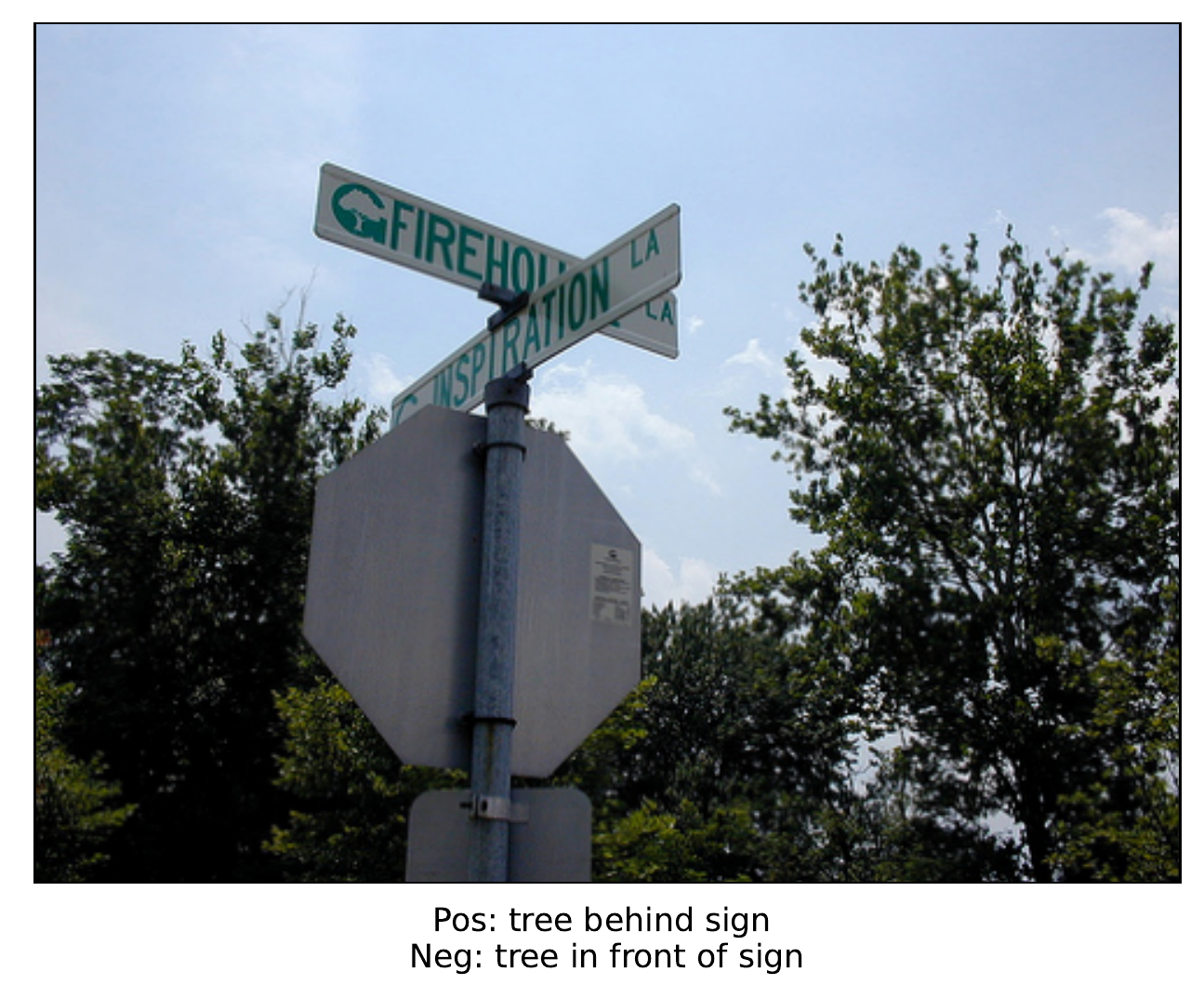}\includegraphics[width=0.3\textwidth]{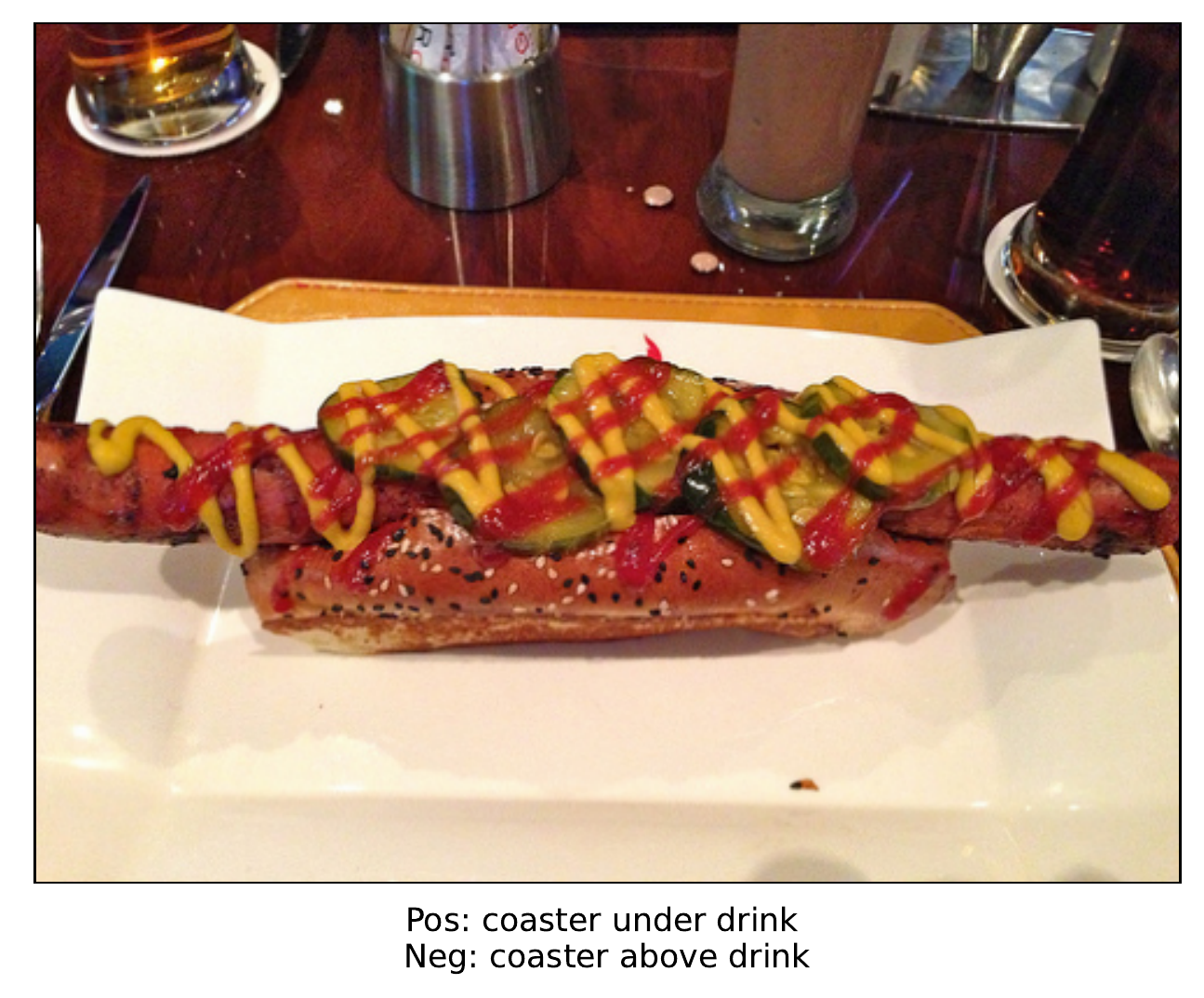}\\

 \end{tabular}
 \caption{Examples where our model correctly chooses the positive caption, while the CLIP baseline fails and incorrectly chooses the negative caption. We show the respective positive and negative captions underneath each image.}
 \label{fig:tf-2}
 \end{figure*}

In this section we qualitatively evaluate the success and errors of our method and that of the CLIP~\cite{clip} baseline. Figures~\ref{fig:tf-1}-\ref{fig:tf-2} show examples where our model succeeds while CLIP model fails.

\begin{figure*}
\includegraphics[width=0.3\textwidth]{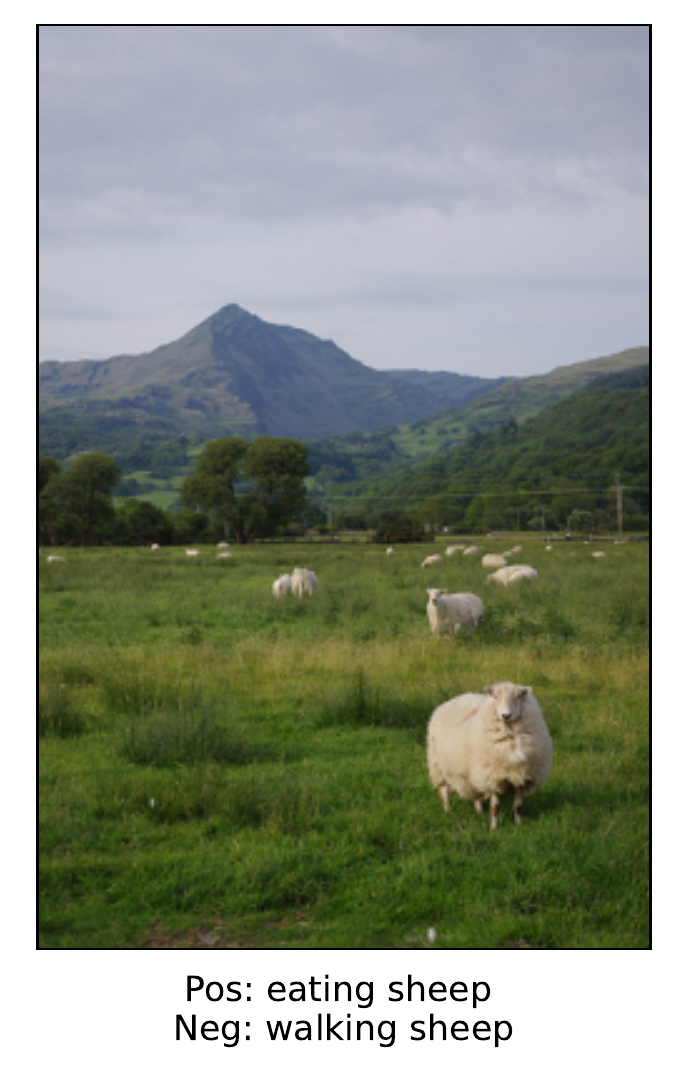}
\includegraphics[width=0.3\textwidth]{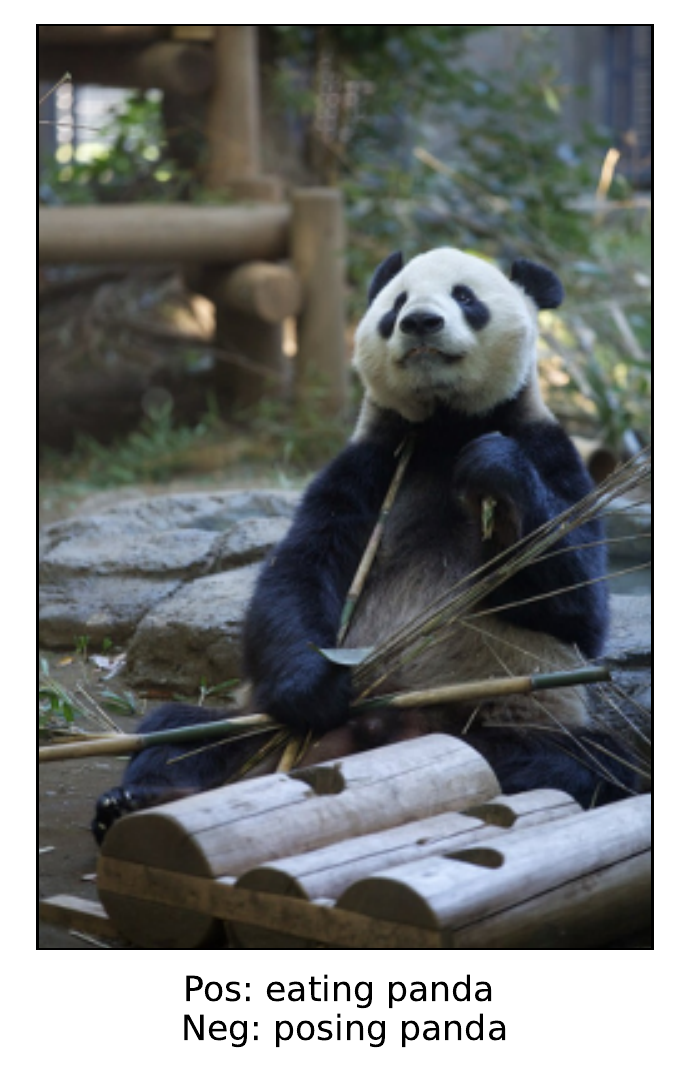}
\includegraphics[width=0.3\textwidth]{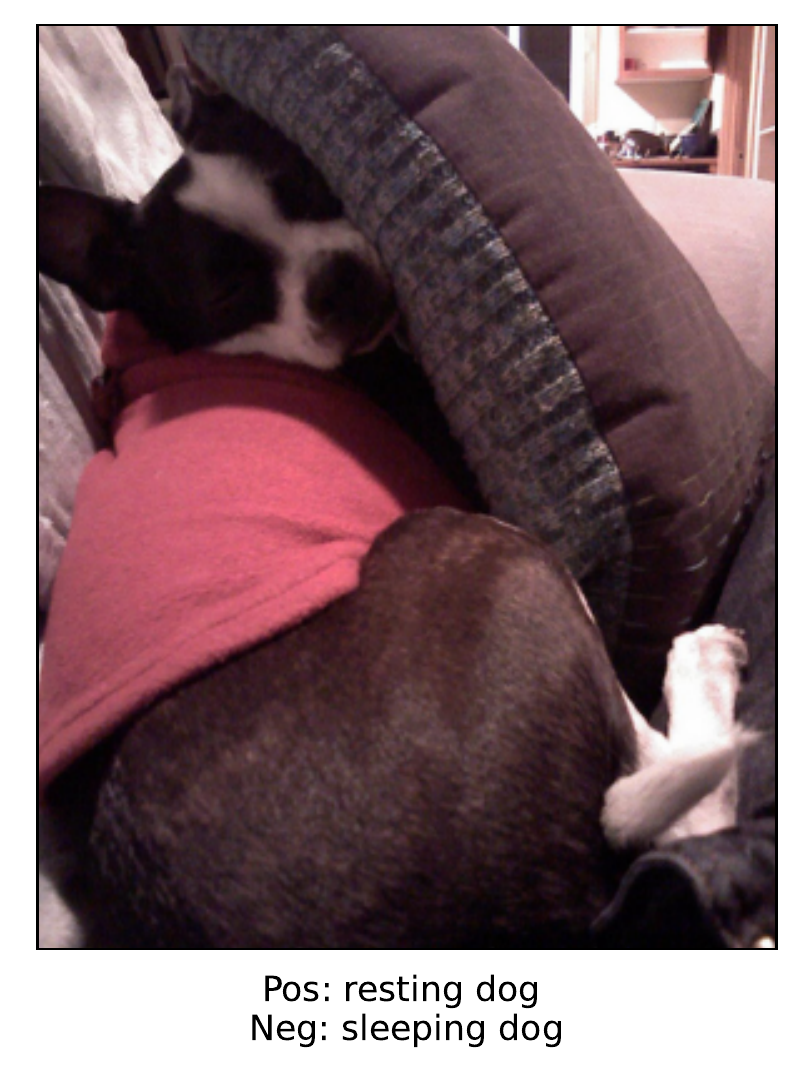}
\\

\includegraphics[width=0.3\textwidth]{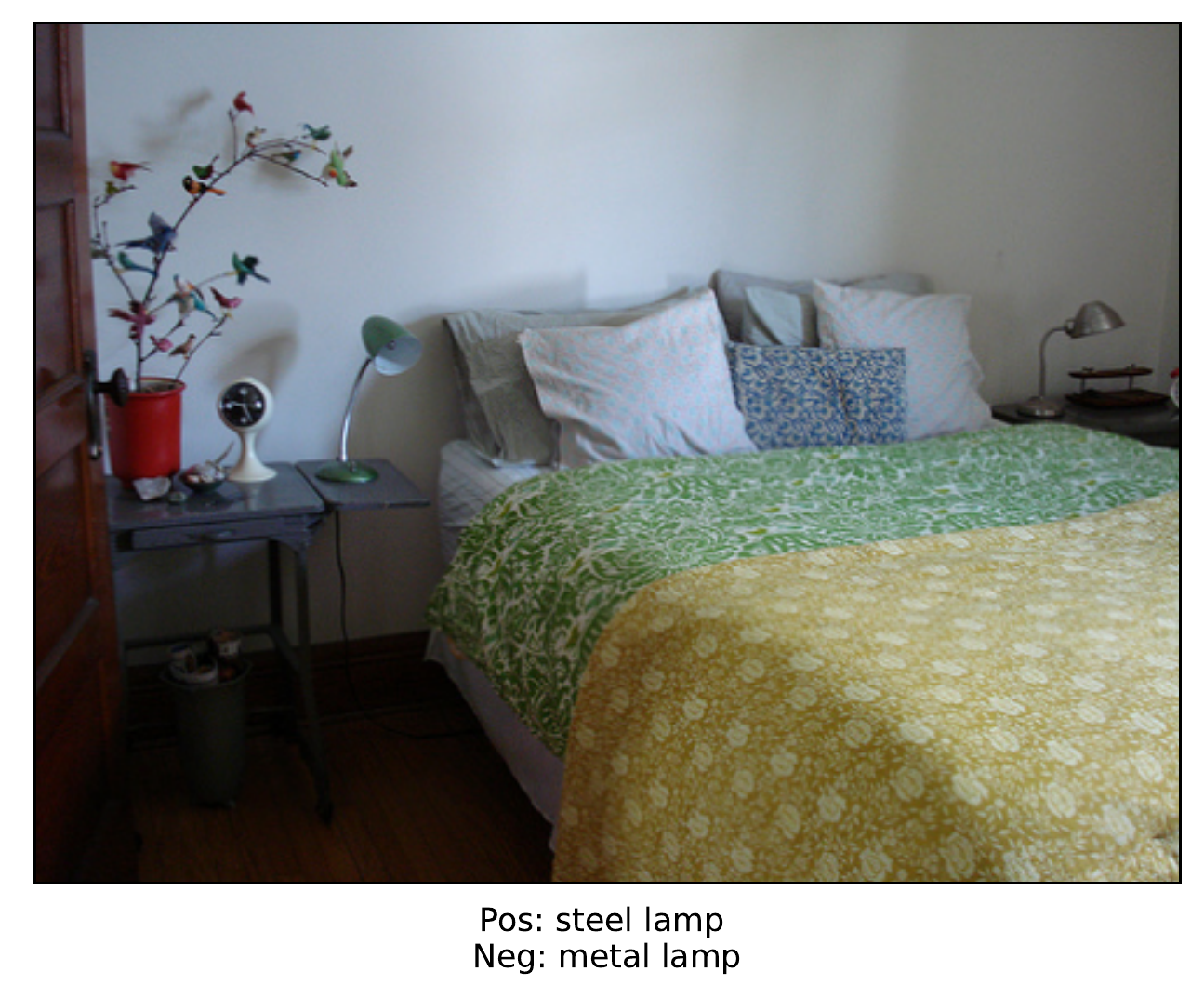}
    \includegraphics[width=0.3\textwidth]{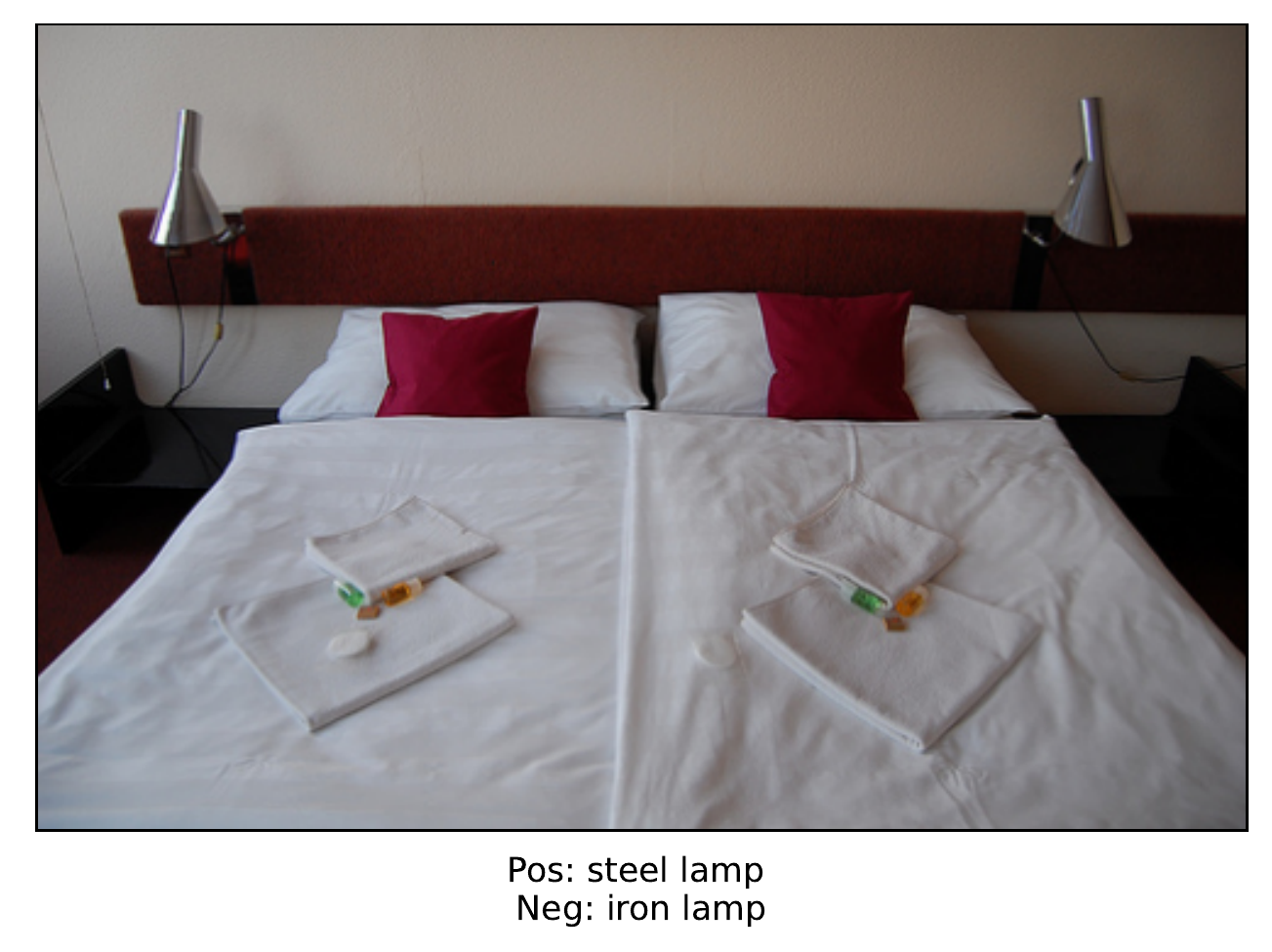}
    \includegraphics[width=0.3\textwidth]{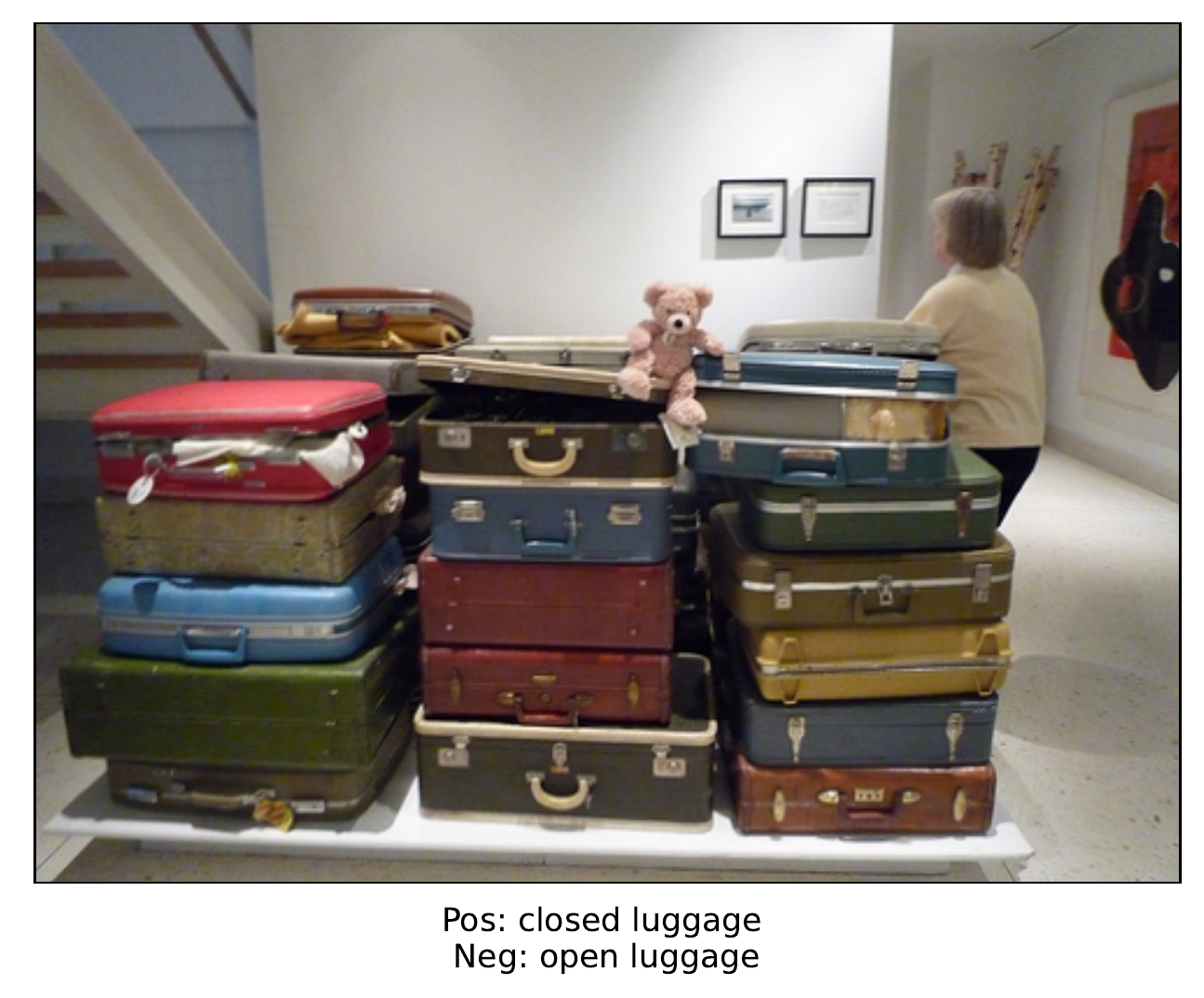}
    \\
\includegraphics[width=0.3\textwidth]{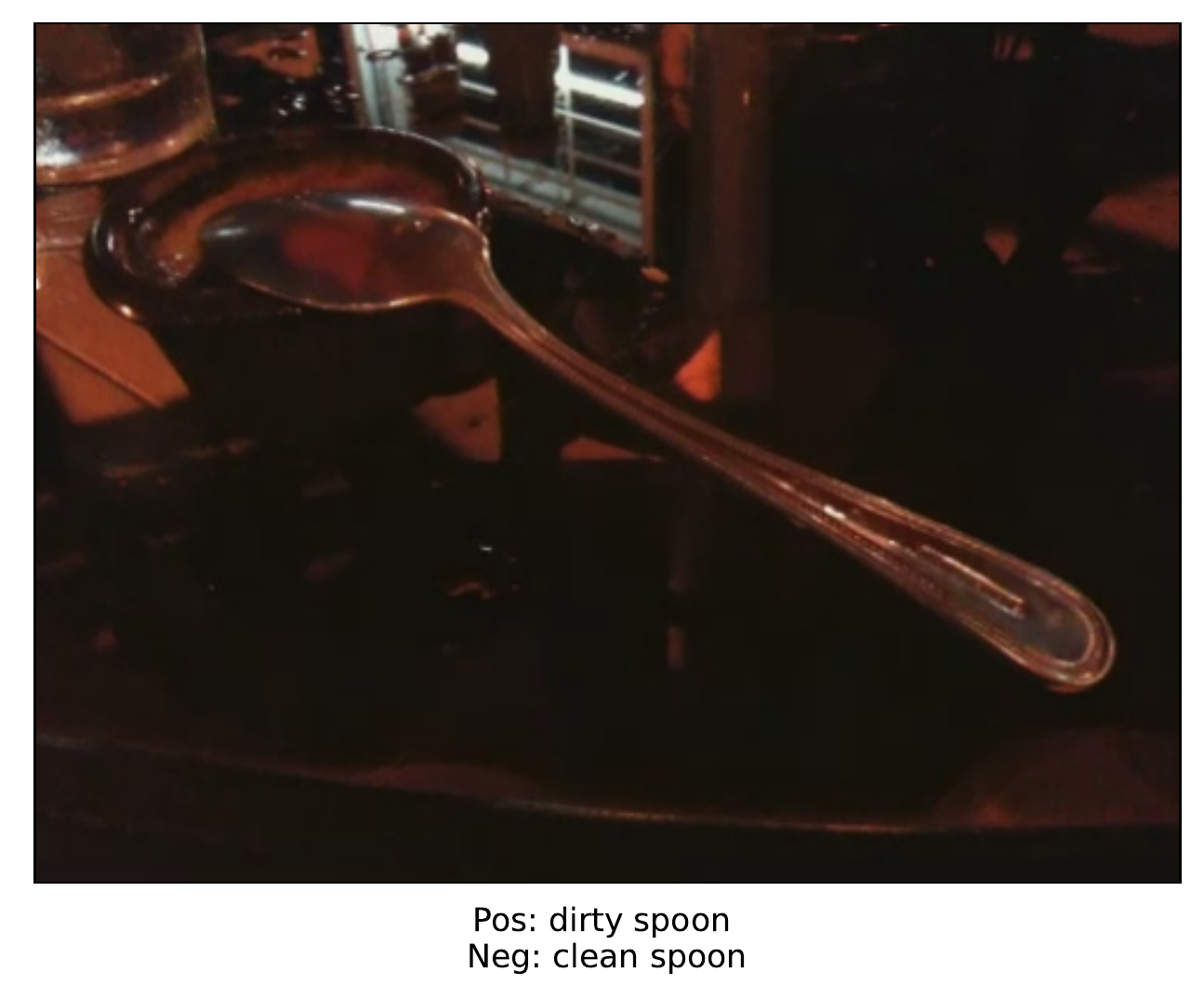}
 \includegraphics[width=0.3\textwidth]{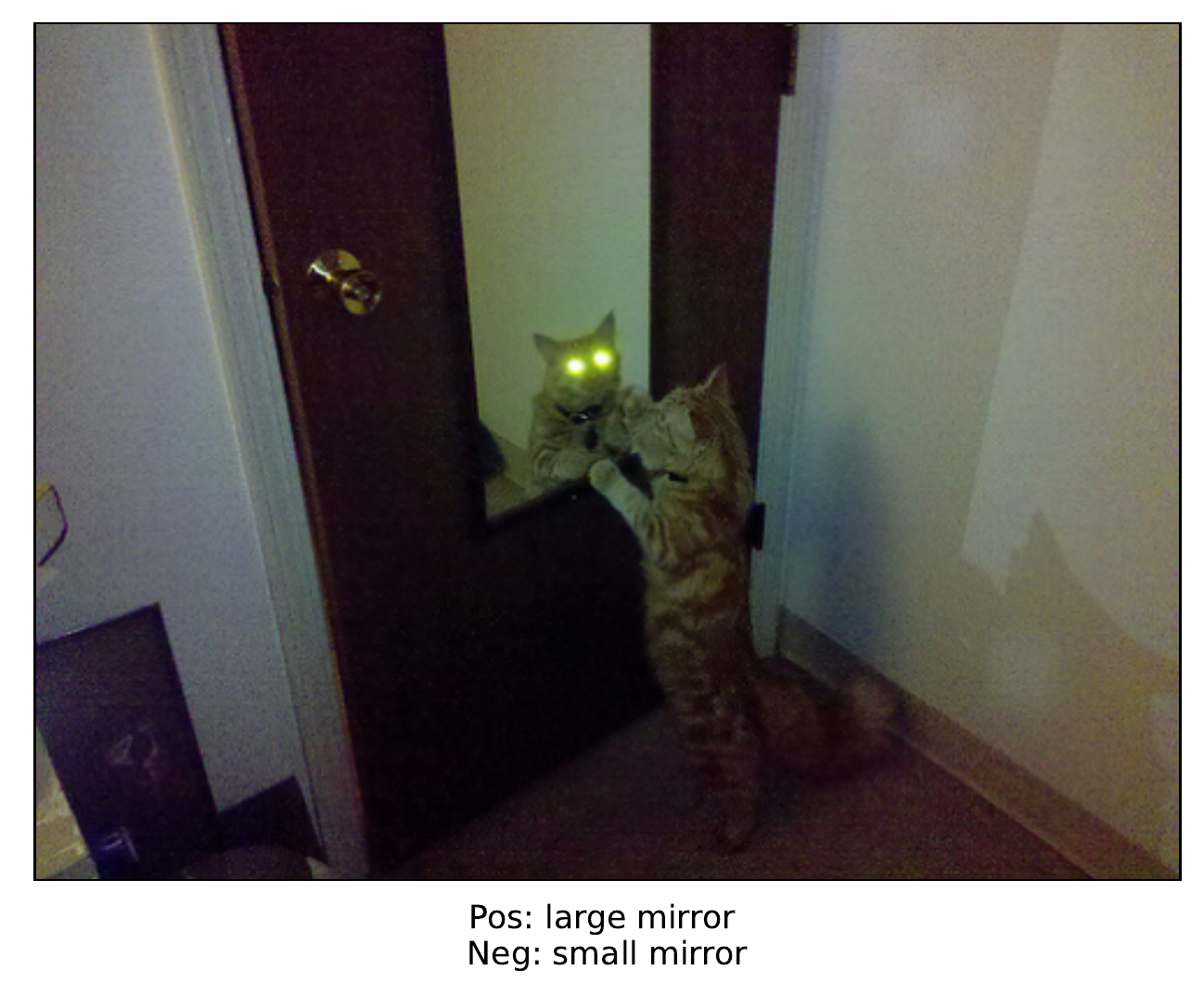}
 \includegraphics[width=0.3\textwidth]{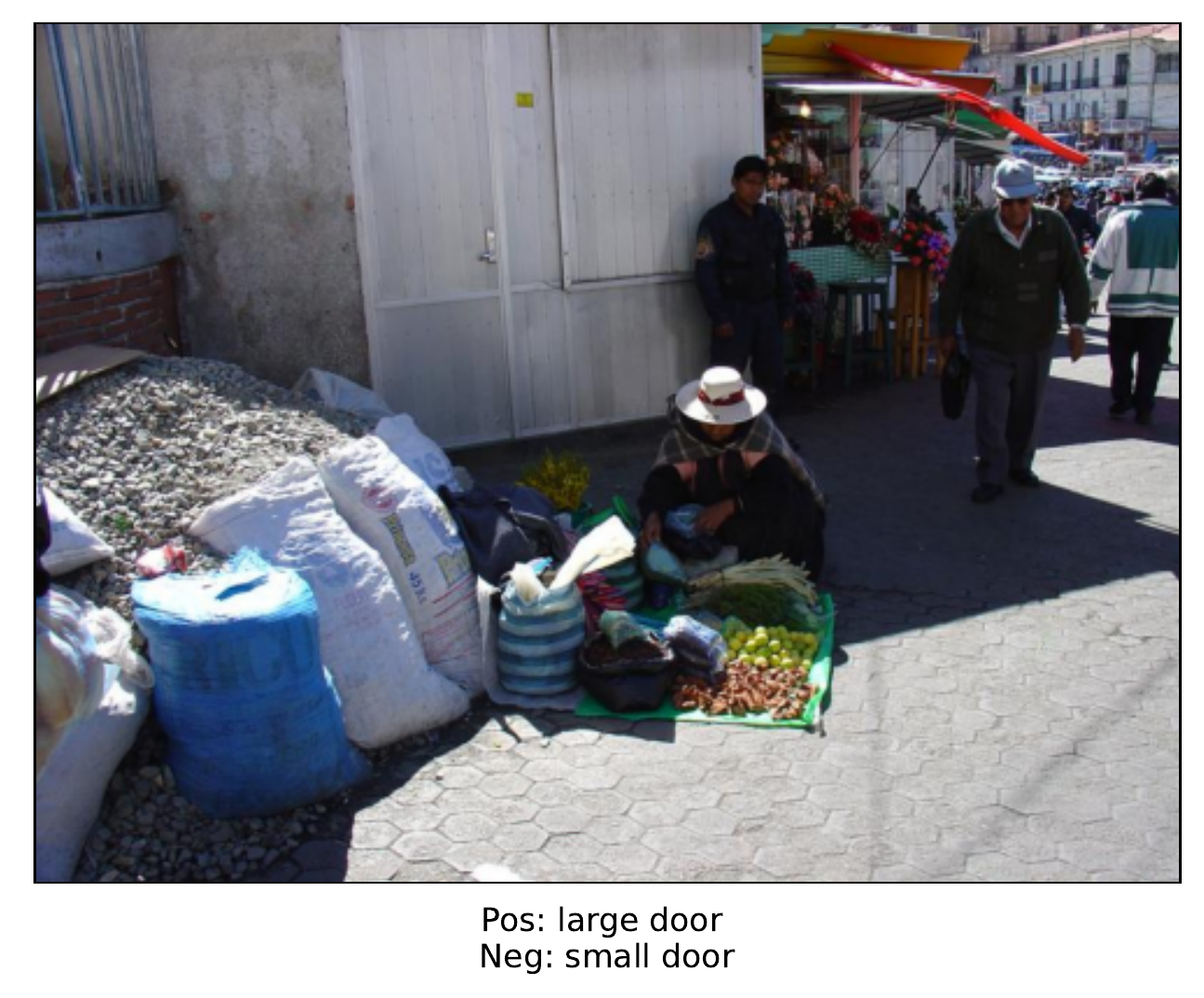}
 \\
 \includegraphics[width=0.3\textwidth]{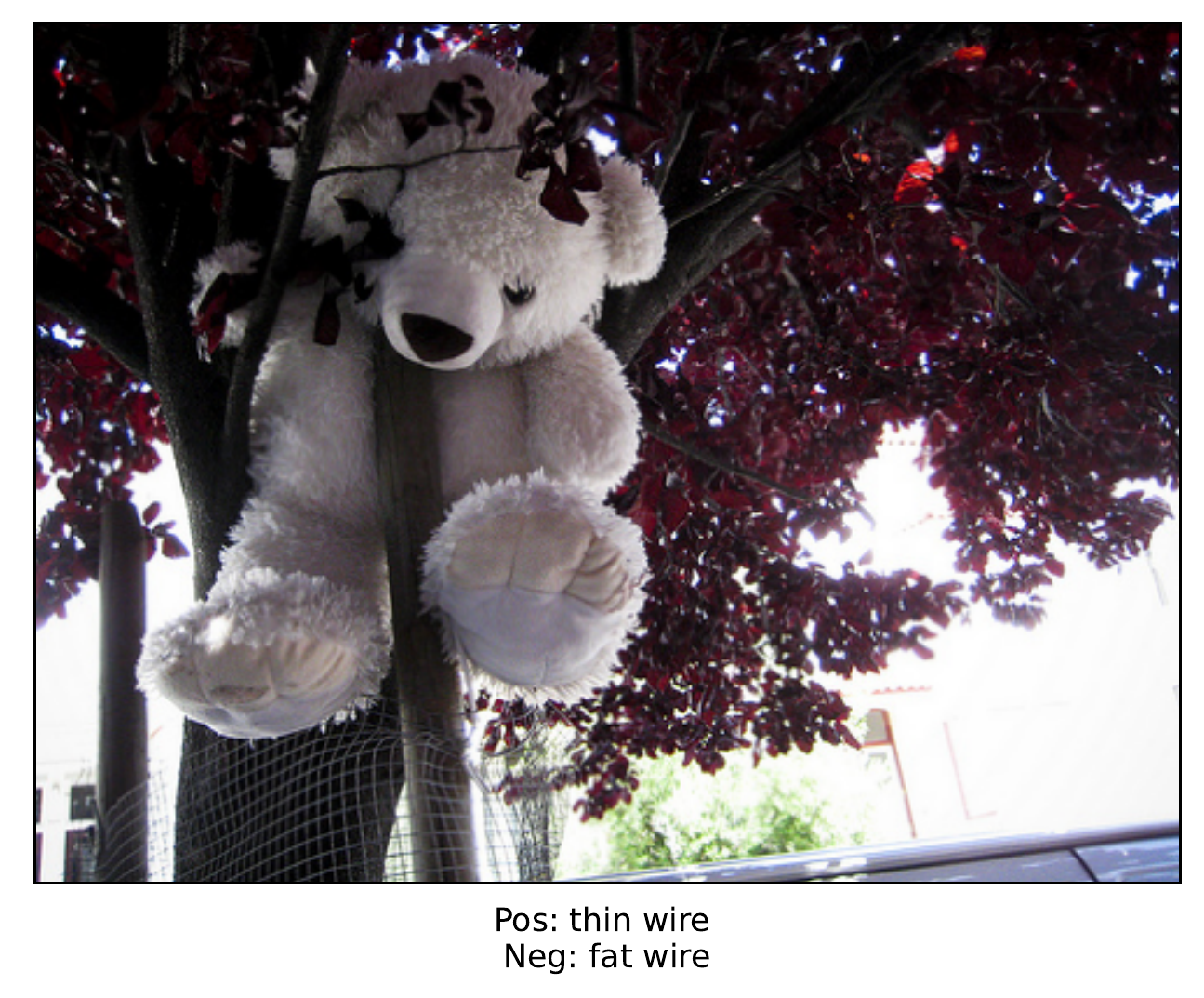}
 \includegraphics[width=0.3\textwidth]{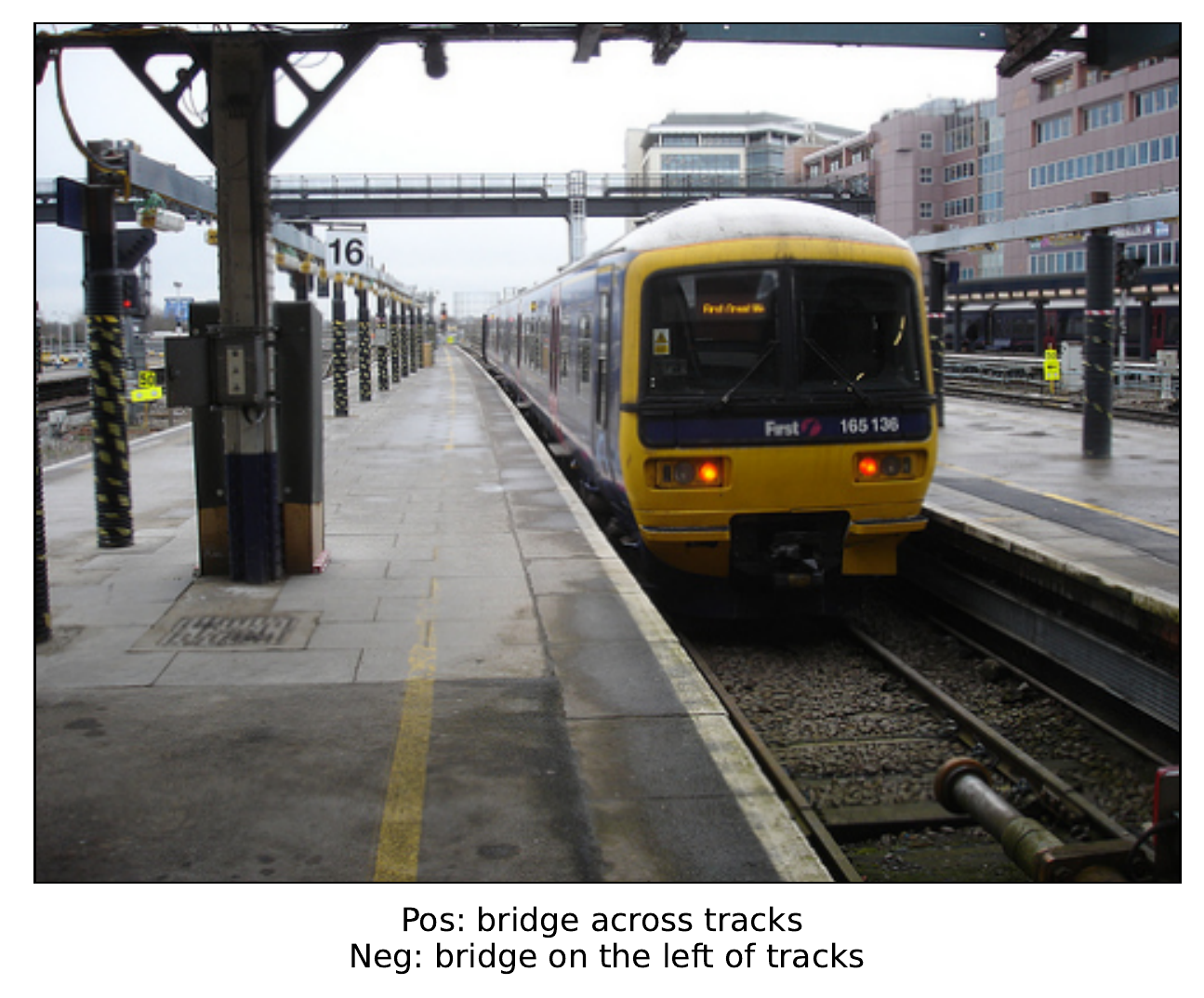}
 \includegraphics[width=0.3\textwidth]{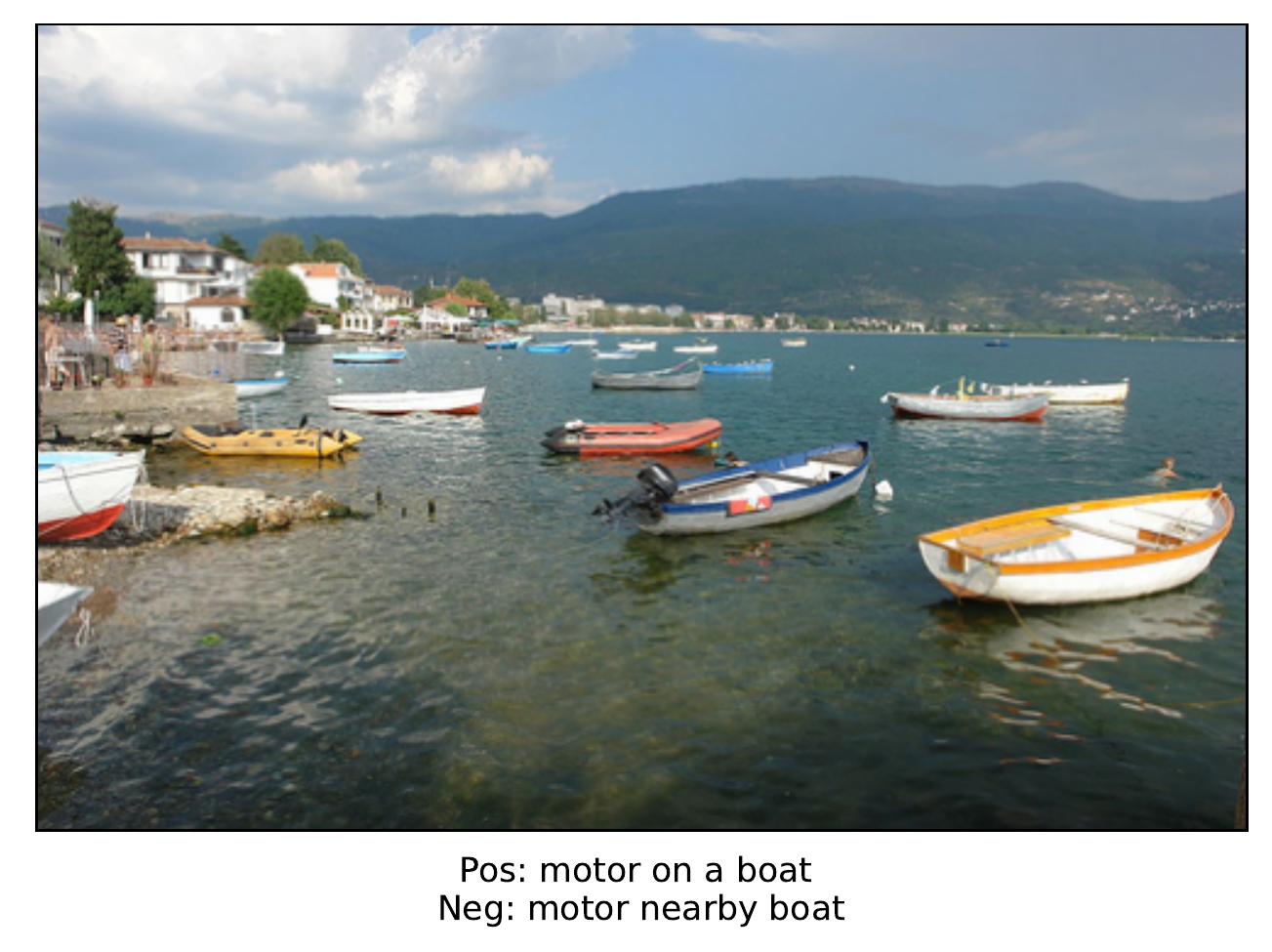}
 \\


    
     
 \caption{Failure cases of our method. In most cases, they are justifiable as both positive and negative captions match the respective images. Notably, these examples are also failing CLIP. }
 \label{fig:ff-2}
 \end{figure*}

Figure \ref{fig:ff-2} show examples of our method failures. It is evident that most of these failure cases are ambiguous even for a human observer. For example, in the second row of figure~\ref{fig:ff-2} there is an image with a lot of suitcases, some are open and some are closed. Therefore, both the positive caption ``closed luggage" and the negative caption ``open luggage" are valid captions in this case.






\section{Code}
Our code and pretrained models are available at: \url{https://github.com/SivanDoveh/TSVLC}


\begin{table*}[]
\centering
\begin{tabular}{l|c|c|c|c|c|c|c}
\toprule
 & O-Large & O-Medium & O-Small & O-Center & O-Mid & O-Margin & Avg O\\ 
 \midrule 
CLIP~\cite{clip} & 86.95 & 77.75 & 72.75 & 85.5 & 80.5 & 70.6 & 79.00\\ 
CLIP +LoRA & 86.5 & 75.9 & 71.65 & 85.25 & 78.25 & 67.25 & 77.46 \\ 
\midrule 
Ours RB+LLM Negs & 91.7 & \textbf{83.2} &\textbf{ 78.9} & \textbf{90.3} & \textbf{84.55} & \textbf{77.4} & \textbf{84.34}\\ 
Ours Combined & \textbf{90.5} & 81.95 & 77.6 & 89.75 & 83.8 & 73.35 & 82.82 \\ 
\bottomrule
\end{tabular}
\caption{Results of \textbf{fine-tuning} on CC3M evaluated on the \textbf{VG} dataset - \textbf{Objects}}
\label{FT_VG_Obj}
\end{table*}

\begin{table*}[]
\centering
\begin{tabular}{l|c|c|c|c|c|c|c|c}
\toprule
 & A-Color & A-Material & A-Size & A-State & A-Action & R-action & R-spatial & Avg A+R \\ 
 \midrule 

CLIP~\cite{clip} & 68.9 & 65.4 & 72.1 & 69.3 & 72.37 & 62.4 & 54 & 66.35 \\ 
CLIP +LoRA & 72.3 & 64.8 & 69.4 & 63.6 & 69.71 & 55.7 & 41 & 62.36 \\ 
\midrule 
Ours RB+LLM Negs & \textbf{82.7} & \textbf{84.9} & \textbf{78.1} & \textbf{71.6} & \textbf{75.13} & \textbf{70.00} & \textbf{78.4} & \textbf{77.26} \\ 
Ours Combined & 79.9 & 78 & 76.8 & 68.7 & 74.18 & 61.9 & 63.2 & 71.81 \\ 
\bottomrule
\end{tabular}
\caption{Results of \textbf{fine-tuning} on CC3M evaluated on the \textbf{VG} dataset - \textbf{Attributes, Relations}}
\end{table*}

\begin{table*}[]
\centering
\begin{tabular}{l|c|c|c|c|c|c|c|c}
\toprule
                      & O-Large & O-Medium & O-Small & O-Center & O-Mid   & O-Margin & R-action & Avg All    \\ 
                      \midrule
CLIP~\cite{clip}                  & 76.975  & 73.28    & 59.41 & 78.075   & 74.63   & 64.49    & 77.2     & 72.00 \\ 
CLIP +LoRA           & 80.82  & 75.02   & 60.81 & 81.6     & 76.94 & 68.37  & 81.8     & 75.05 \\ 
\midrule 
Ours RB+LLM Negs      & 82.77  & 77.97  & 67.34 & 83.05    & 79.92 & \textbf{75.36}    & \textbf{87.6}     & 79.14\\ 
Ours Combined & \textbf{83.5}    & \textbf{80.05}    & \textbf{71.70} & \textbf{84.02}   & \textbf{81.17}   & 75.01    & 84.2     & \textbf{79.95} \\ \bottomrule
\end{tabular}
\caption{Results of \textbf{fine-tuning} on CC3M evaluated on the \textbf{SWIG} dataset}
\end{table*}

\begin{table*}[]
\centering
\begin{tabular}{l|c|c|c|c|c|c|c|c}
\toprule
                      & O-Large & O-Medium & O-Small & O-Center & O-Mid & O-Margin & R-action & Avg All    \\ \midrule
CLIP~\cite{clip}                  & \textbf{97.9}    & \textbf{93.3}     & \textbf{90.00}      & 98.6     & 98.1  & \textbf{89.7 }    & 78.2     & \textbf{92.25} \\ 
CLIP +LoRA            & 96.2    & 89.2     & 85.1    & 97.7     & 96.4  & 83.7     & 72.6     & 88.7       \\ 
\midrule 
Ours RB+LLM Negs     & \textbf{97.9}    & 89.8     & 88.4    & \textbf{99.2}     & 97.7  & 86.1     & \textbf{79.4}     & 91.21 \\ 
Ours Combined & 97.6    & 89.8     & 86.5    & 98.6     & \textbf{98.5}  & 86.6     & 78       & 90.8       \\ \bottomrule
\end{tabular}
\caption{Results of \textbf{fine-tuning} on CC3M evaluated on the \textbf{HAKE} dataset}
\end{table*}

\begin{table*}[]
\centering
\begin{tabular}{l|c|c|c|c|c|c}
\toprule
                      & A-Color & A-Material & A-Size & A-State & A-Action & Avg All \\ \midrule
CLIP~\cite{clip}                 & 71      & 73.3       & 68     & 53.3    & 62.7     & 65.66   \\ 
CLIP +LoRA           & 74      & 71.4       & 66.9   & 51.6    & 59.1     & 64.6    \\ 
\midrule
Ours RB+LLM Negs      & \textbf{75.7 }   & \textbf{83.5}       & 66.3   & \textbf{56.6}    & \textbf{64.6}     & \textbf{69.34}   \\ 
Ours Combined & 75      & 76.7       & \textbf{69.9 }  & 55.9    & \textbf{64.6}     & 68.42   \\ \bottomrule
\end{tabular}
\caption{Results of \textbf{fine-tuning} on CC3M evaluated on the \textbf{VAW} dataset}
\label{FT-Vaw}
\end{table*}

\begin{table*}[]
\centering
\begin{tabular}{l|c|c|c|c|c|c|c}
\toprule 
 & O-Large & O-Medium & O-Small & O-Center & O-Mid & O-Margin & Avg O \\ \midrule
CLIP~\cite{clip} & 76.5 & 66.15 & \textbf{64.35} & 74.85 & 66.6 & 62.35 & 68.46 \\ 
\midrule 
Ours RB+LLM Negs & \textbf{79.4 }& \textbf{67.8} & 62.15 & \textbf{75.15} & \textbf{70.00} & \textbf{64.7} & \textbf{69.86}\\
Ours Combined & 76.7 & 67.4 & 62.15 & 74.7 & 67.85 & 64.15 & 68.82\\ 
\bottomrule
\end{tabular}
\caption{Results of \textbf{training from scratch} on CC3M evaluated on the \textbf{VG} dataset - \textbf{Objects}}
\label{FS_VG_obj}
\end{table*}

\begin{table*}[]
\centering
\begin{tabular}{l|c|c|c|c|c|c|c|c}
\toprule

 & A-Color & A-Material & A-Size & A-State & A-Action & R-action & R-spatial & Avg A+R \\ \midrule
CLIP~\cite{clip} & 62 & 58.3 & \textbf{68.4} & 46.8 & 63.87 & 44.3 & 32.5 & 53.73 \\ 
\midrule 
Ours RB+LLM Negs  & 72.4 & \textbf{74.4} & 57 & \textbf{61.2} & \textbf{75.35} & \textbf{54.7} &\textbf{ 82.3} &\textbf{ 68.19} \\
Ours Combined & \textbf{74} & 64.4 & 65.9 & 54.6 & 70.99 & 51.4 & 56.2 & 62.49 \\ \bottomrule
\end{tabular}
\caption{Results of \textbf{training from scratch} on CC3M evaluated on the \textbf{VG} dataset - \textbf{Attributes, Relations}}
\label{FS_VG}
\end{table*}

\begin{table*}[]
\centering
\begin{tabular}{l|c|c|c|c|c|c|c|c}
\toprule
                          & O-Large & O-Medium & O-Small & O-Center & O-Mid   & O-Margin & R-action & Avg All    \\ \midrule
CLIP~\cite{clip}       & \textbf{68.15 }  & 62.36   & 58.60 & \textbf{68.15}    & 63.74 & 60.26  &\textbf{ 65.9}     & 63.88 \\ 
\midrule 
Ours RB+LLM Negs      & 66.02  & 63.60  &\textbf{ 61.51 }& 65.92   & 64.10 & 63.94  & 61.4     & 63.78 \\ 
Ours Combined & 67.57  &\textbf{ 64.94}  & 60.21 & 66.55    & \textbf{65.52} & \textbf{67.49}    & 60.4     & \textbf{64.67} \\ 
\bottomrule
\end{tabular}
\caption{Results of \textbf{training from scratch} on CC3M evaluated on the \textbf{SWIG} dataset}
\end{table*}

\begin{table*}[]
\centering
\begin{tabular}{l|c|c|c|c|c|c|c|c}
\toprule
                          & O-Large & O-Medium & O-Small & O-Center & O-Mid & O-Margin & R-action & Avg All    \\ \midrule
CLIP~\cite{clip}        & 87.7    & 78.8     & 72      & 90.4     & 85    & 75.1     & 63.5     & 78.92 \\ 
\midrule 

Ours RB+LLM Negs      & 86.8    & \textbf{85.8}     & \textbf{79.6}    & \textbf{91.9 }    & 86    & \textbf{79.2 }    & \textbf{74.8}     & \textbf{83.44 }\\ 
Ours Combined & \textbf{88.1}    & 76.5     & 72.7    & 90.5     & \textbf{86.1}  & 73       & 68.3     & 79.31 \\ \bottomrule
\end{tabular}
\caption{Results of \textbf{training from scratch} on CC3M evaluated on the \textbf{HAKE} dataset}
\end{table*}

\begin{table*}[]
\centering
\begin{tabular}{l|c|c|c|c|c|c}
\toprule
                          & A-Color & A-Material & A-Size & A-State & A-Action & Avg All \\ \midrule
CLIP~\cite{clip}        & 55.4    & 57         & 61.3   & 48.3    & 57       & 55.8    \\ 
\midrule 
Ours RB+LLM Negs      & \textbf{75.1}    &\textbf{ 68.6}       & 56.3   & \textbf{61.1}    & \textbf{60.2}     & \textbf{64.26 }  \\ 
Ours Combined & 61.3    & 63.7       & \textbf{68.6}   & 53.8    & 55.5     & 60.58   \\ \bottomrule
\end{tabular}
\caption{Results of \textbf{training from scratch} on CC3M evaluated on the \textbf{VAW} dataset}
\label{FS_VAW_Att}
\end{table*}

\clearpage

\clearpage
{\small
\bibliographystyle{ieee_fullname}
\bibliography{egbib}
}